\documentclass[journal]{IEEEtran}
\usepackage{cite}

\ifCLASSINFOpdf
\usepackage[pdftex]{graphicx}
\graphicspath{{/}}
\else
\usepackage[dvips]{graphicx}
\graphicspath{{figures/}}
\fi

\usepackage{amsmath}
\usepackage{array}

\usepackage{url}

\usepackage{balance}
\usepackage{jmei_custom}

\begin{document}
%
\title{SILVar: Single Index Latent Variable Models}
\author{Jonathan Mei
	and~Jos\'{e}~M.F.~Moura
	\thanks{This work partially funded by NSF grants CCF 1011903 and CCF
		1513936.}
	\thanks{This paper appears in: IEEE Transactions on Signal Processing, online March 21, 2018}
\thanks{Print ISSN: 1053-587X,  Online ISSN: 1941-0476}
\thanks{Digital Object Identifier: 10.1109/TSP.2018.2818075}}

\markboth{IEEE Transactions on Signal Processing}%
{Mei \MakeLowercase{\textit{et al.}}: SILVar}


\maketitle

\begin{abstract}
	A semi-parametric, non-linear regression model in the presence of latent variables is introduced. These latent variables can correspond to unmodeled phenomena or unmeasured agents in a complex networked system. This new formulation allows joint estimation of certain non-linearities in the system, the direct interactions between measured variables, and the effects of unmodeled elements on the observed system. The particular form of the model adopted is justified, and learning is posed as a regularized empirical risk minimization. This leads to classes of structured convex optimization problems with a ``sparse plus low-rank'' flavor. Relations between the proposed model and several common model paradigms, such as those of Robust Principal Component Analysis (PCA) and Vector Autoregression (VAR), are established. Particularly in the VAR setting, the low-rank contributions can come from broad trends exhibited in the time series. Details of the algorithm for learning the model are presented. Experiments demonstrate the performance of the model and the estimation algorithm on simulated and real data.
\end{abstract}
\section{Introduction}

How real is this relationship? This is a ubiquitous question that presents itself not only in judging interpersonal connections but also in evaluating correlations and causality throughout science and engineering. Two reasons for reaching incorrect conclusions based on observed relationships in collected data are chance and outside influences. For example, we can flip two coins that both show heads, or observe that today's temperature measurements on the west coast of the continental USA seem to correlate with tomorrow's on the east coast throughout the year. In the first case, we might not immediately conclude that coins are related, since the number of flips we observe is not very large relative to the possible variance of the process, and the apparent link we observed is up to chance. In the second case, we still may hesitate to use west coast weather to understand and predict east coast weather, since in reality both are closely following a seasonal trend.

Establishing interpretable relationships between entities while mitigating the effects of chance can be achieved via sparse optimization methods, such as regression (Lasso)~\cite{tibshirani_regression_1996} and inverse covariance estimation~\cite{friedman_sparse_2008}. In addition, the extension to time series via vector autoregression~\cite{bolstad_causal_2011,basu_regularized_2015} yields interpretations related to Granger causality~\cite{granger_investigating_1969}. In each of these settings, estimated nonzero values correspond to actual relations, while zeros correspond to absence of relations.

However, we are often unable to collect data to observe all relevant variables, and this leads to observing relationships that may be caused by common links with those unobserved variables. The hidden variables in this model are fairly general; they can possibly model underlying trends in the data, or the effects of a larger network on an observed subnetwork. For example, one year of daily temperature measurements across a country could be related through a graph based on geographical and meteorological features, but all exhibit the same significant trend due to the changing seasons. We have no single sensor that directly measures this trend. In the literature, a standard pipeline is to de-trend the data as a preprocessing step, and then estimate or use a graph to describe the variations of the data on top of the underlying trends~\cite{sandryhaila_discrete_2013,sandryhaila_discrete_2014,mei_signal_2017}.

Alternatively, attempts have been made to capture the effects of hidden variables via sparse plus low-rank optimization~\cite{chandrasekaran_latent_2012}. This has been extended to time series~\cite{jalali_learning_2011}, and even to a non-linear setting via Generalized Linear Models (GLMs)~\cite{bahadori_fast_2013}. What if the form of the non-linearity (link function) is not known? Regression using a GLM with an unknown link function is also known as a Single Index Model (SIM). Recent results have shown good performance when using SIMs for sparse regression~\cite{ganti_learning_2015}.

So far, when choosing a model, current methods will impose a fixed form for the (non-)linearity, assume the absence of any underlying trends, perform separate pre-processing or partitioning in an attempt to remove or otherwise explicitly handle such trends, or take some combination of these steps.
To address all of these issues, we propose a model with a non-linear function applied to a linear argument that captures the effects of latent variables, which manifest as unmodeled trends in the data. Thus, we introduce the Single Index Latent Variable (SILVar) model, which uses the SIM in a sparse plus low-rank  optimization setting to enable general, interpretable multi-task regression in the presence of unknown non-linearities and unobserved variables. That is, we propose the SILVar model not only to use for regression in difficult settings, but also as a tool for uncovering hidden relationships buried in data.

First, we establish notation and review prerequisites in Section~\ref{sec:bg}. Next, we introduce the SILVar model and discuss several paradigms in which it can be applied in Section~\ref{sec:SILVar}. Then, we detail the numerical procedure for learning the SILVar model in Section~\ref{sec:learn_SILVar}. Finally, we demonstrate the performance via experiments on synthetic and real data in Section~\ref{sec:exp}.

\section{Background}
\label{sec:bg}
In this section, we introduce the background concepts and notation used throughout the remainder of the paper.
\subsection{Bregman Divergence and Convex Conjugates}
For a given convex function $\phi$, the Bregman Divergence~\cite{bregman_relaxation_1967} associated with $\phi$ between $\y$ and $\x$ is denoted
\begin{equation}
\label{eq:breg_div}
D_\phi(\y \| \x ) = \phi(\y) - \phi(\x) - \nabla\phi(\x)^\top (\y-\x).
\end{equation}
The Bregman Divergence is a non-negative (asymmetric) quasi-metric. Two familiar special cases of the Bregman Divergence are Euclidean Distance when $\phi(\x)=\frac{1}{2}\|\x\|_2^2$ and Kullback-Liebler Divergence when $\phi(\x)=\sum\limits_i x_i\log x_i$ in the case that $\x$ is a valid probability distribution (i.e., $\x\ge 0$ and $\sum\limits_i x_i = 1$).

The convex conjugate $\phi_*$ of a function $\phi$ is given by
\begin{equation}
\label{eq:conv_conj}
\phi_*(\x) \overset{\Delta}{=} \sup_\y \; \y^\top\x-\phi(\y).
\end{equation}
The convex conjugate arises naturally in optimization problems when deriving a dual form for the original (primal) problem. For closed, convex, differentiable, 1-D function $\phi$ with invertible gradient, the following properties hold
\begin{equation}
\label{eq:conv_conj_prop}
\begin{aligned}
&\phi_*(x)=x(\nabla\phi)^{-1}(x)-\phi((\nabla\phi)^{-1}(x)) \\
&(\nabla \phi)^{-1} =\nabla \phi_* \qquad\qquad (\phi_*)_* =\phi
\end{aligned}
\end{equation}
where $(\cdot)^{-1}$ denotes the inverse function, not the multiplicative inverse. In words, these properties give an alternate form of the conjugate in terms of gradients of the original function, state that the function inverse of the gradient is equal to the gradient of the conjugate, and state that the conjugate of the conjugate is the original function.

\subsection{Generalized Linear and Single-Index Models}
\label{sec:bg:GLM_SIM}
The Generalized Linear Model (GLM) can be described using several parameterizations. We adopt the one based on the Bregman Divergence~\cite{banerjee_clustering_2005}. For observations $y_i\in\Rbb$ and $\x_i\in\Rbb^{p}$, let $\y=(y_1\ldots y_n)^\top$, $\X=(\x_1\ldots\x_n)$. The model is parameterized by 1) a non-linear link function $g=(\nabla \phi)^{-1}$ where  $\phi$ is a closed, convex, differentiable, invertible function; and 2) a vector $\a\in\Rbb^p$. We have the model written as
\begin{equation}
\label{eq:GLM_expec}
\Ebb[y_i | \x_i] = g \left(\a^\top \x_i \right),
\end{equation}
(note that some references use $g^{-1}$ as the link function where we use $g$) and the corresponding likelihood function written as
\begin{equation}
\label{eq:GLM}
P(y_i | \x_i) = \exp \left\{-D_\phi \left(y_i \| g \left(\a^\top \x_i \right) \right) \right\}
\end{equation}
where the likelihood is expressed with respect to an appropriate base measure~\cite{acharyya_parameter_2015}, which can be omitted for notational clarity.	

Let $G=\phi_*$ and $g=\nabla G=(\nabla\phi)^{-1}$. Then, for data $\{\x_i,y_i\}$ with conditionally independent $y_i$ given $\x_i$ (note that this is not necessarily assuming that $\x_i$ are independent), learning the model $\a$ assuming $g$ is known can be achieved via empirical risk minimization,
\begin{equation}
\label{eq:GLM_MLE}
\begin{aligned}
\what{\a} &= \underset{\a}{\arg\!\max}\; \prod\limits_{i=1}^{n}\; \exp \left\{-D_\phi \left(y_i \| g \left(\a^\top \x_i\right) \right) \right\}\\
&= \argmin[\a] \sum\limits_{i=1}^{n}\;D_\phi \left(y_i \| g \left(\a^\top \x_i \right) \right)\\
&=\argmin[\a] \sum\limits_{i=1}^{n} \begin{aligned}[t]&\left[ \phi\left(y_i\right)- \phi\left(g\left(\a^\top\x_i\right)\right) \right.\\
&\quad \left. -\nabla\phi\left(g\left(\a^\top\x_i\right)\right)\left(y_i - g\left(\a^\top\x_i\right)\right) \right]\end{aligned}\\
&\overset{(a)}{=}\argmin[\a] \sum\limits_{i=1}^{n} \begin{aligned}[t]&\left[ G_*\left(y_i\right) - y_i \left(\a^\top \x_i\right) - \phi\left(g\left(\a^\top\x_i\right)\right) \right.\\
&\quad \left. +\left(\a^\top\x_i\right) g\left(\a^\top\x_i\right) \right]\end{aligned}\\
&\overset{(b)}{=} \argmin[\a] \sum\limits_{i=1}^{n}\;\left[ G_*\left(y_i\right) + G\left(\a^\top \x_i\right) - y_i \left(\a^\top \x_i\right) \right]\\
&= \argmin[\a] \what{F}_1\left(\y,\X,g,\a\right)
\end{aligned}
\end{equation}
where equality $(a)$ arises from the second property in~\eqref{eq:conv_conj_prop}, equality $(b)$ arises from the first property, and we introduce
{\small
	\begin{equation}
	\label{eq:SIM_MLE_conv_obj}
	\what{F}_1(\y,\X,g,\a) \overset{\Delta}{=}\frac{1}{n}\sum\limits_{i=1}^{n}\;\left[ G_*(y_i) + G\left(\a^\top \x_i \right) - y_i \left(\a^\top \x_i \right) \right]
	\end{equation}
}for notational compactness.

The Single Index Model (SIM)~\cite{ichimura_semiparametric_1993} takes the same form as the GLM. The crucial difference is in the estimation of the models. When learning a GLM, the link function $g$ is known and the linear parameter $\a$ is estimated; however when learning a SIM, the link function $g$ needs to be estimated along with the linear parameter $\a$.

Recently, it has been shown that, when the function $g$ is restricted to be monotonic increasing and Lipschitz, learning SIMs becomes computationally tractable~\cite{acharyya_parameter_2015} with performance guarantees in high-dimensional settings~\cite{ganti_learning_2015}. Thus, with scalar $u$ defining the set $\Ccal^{u}=\{g : \forall y>x,\; 0 \le g(y)-g(x) \le u(y-x) \}$ of monotonic increasing $u$-Lipschitz functions, this leads to the optimization problem,
\begin{equation}
\label{eq:SIM_MLE_conv}
\begin{aligned}
(\what{g},\what{\a}) &= \argmin[g, \a] \what{F}_1(\y,\X,g,\a) \\
& \qquad \textrm{s.t. } g=\nabla G \in\Ccal^{1}.
\end{aligned}
\end{equation}

\subsection{Lipschitz Monotonic Regression and Estimating SIMs}
\label{sec:bg:LMR}
The estimation of $g$ with the objective function including terms $G$ and $G_*$ at first appears to be an intractably difficult calculus of variations problem. However, there is a marginalization technique that cleverly avoids directly estimating functional gradients with respect to $G$ and $G_*$~\cite{acharyya_parameter_2015} and admits gradient-based optimization algorithms for learning. The marginalization utilizes Lipschitz monotonic regression (LMR) as a subproblem. Thus, before introducing this marginalization, we first review LMR.
\subsubsection{LMR}
\label{sec:bg:LMR:LMR}
Given ordered pairs $\{x_i,y_i\}$ and independent Gaussian $w_i$, consider the model
\begin{equation}
y_i = g(x_i) + w_i,
\end{equation}
which intuitively treats $\{y_i\}$ as noisy observations of a function $g$ indexed by $x$, sampled at points $\{x_i\}$. Let $\what{g}_i=g(x_i)$, an estimate of the function value, with $\what{\g}=(\what{g}_1 \ldots \what{g}_n)^\top$, and $x_{[j]}$ denote the $j^{th}$ element of the $\{x_i\}$ sorted in ascending order. Then LMR is described by the problem,
{\small
	\begin{equation}
	\label{eq:Lip_mon_reg}
	\begin{aligned}
	\what{\g} \overset{\Delta}{=} \textrm{LMR}(\y,\x) &= \argmin[\g] \sum\limits_{i=1}^{n}\;\left( g(x_i) - y_i \right)^2\\
	& \textrm{s.t. } 0 \le g\left(x_{[j+1]}\right) \!-\! g\left(x_{[j]}\right) \le x_{[j+1]} \!-\! x_{[j]} \\
	&\qquad\qquad \textrm{for } j=1,\ldots, n-1.
	\end{aligned}
	\end{equation}
}While there may be in fact infinitely many possible (continuous) monotonic increasing Lipschitz functions $g$ that pass through the points $\what{\g}$, the solution vector $\what{\g}$ is unique. We will introduce a simple yet effective algorithm for solving this problem later in Section~\ref{sec:learn_SILVar:LMR}.
\subsubsection{Estimating SIMs}
\label{sec:bg:LMR:SIM}
We now return to the objective function~\eqref{eq:SIM_MLE_conv}. Let $\what{\g}=\textrm{LMR}(\y,\X^\top\a)$. Then the gradient w.r.t. $\a$ can be expressed in terms of an estimate of $g$ without explicit computation of $G$ or $G_*$,
\begin{equation}
\label{eq:SIM_MLE_marg}
\begin{aligned}
\nabla_\a F_1 &= \sum\limits_{i=1}^{n}\;\left[ \left( \what{g}_i - y_i \right) \x_i \right].
\end{aligned}
\end{equation}
This allows us to apply gradient or quasi-Newton methods to solve the minimization in $\a$, which is itself a convex problem since the original problem was jointly convex in $g$ and $\a$.	

\section{Single Index Latent Variable Models}
\label{sec:SILVar}
In this section, we build the Single Index Latent Variable (SILVar) model from fundamental concepts.

\subsection{Multitask regression and latent variables}
\label{sec:SILVar:multi_reg_lat_var}
First, we extend the SIM to the multivariate case and then examine how latent variables can affect learning of the linear parameter. Let $\y_i=\left(y_{1i} \; \ldots \; y_{mi}
\right)^\top$, $\g(\x)=\left(g_1(x_1) \; \ldots \; g_m(x_m) \right)^\top$, and $\A=\left(\a_1 \; \ldots \; \a_m\right)^\top$. Consider the vectorization,
\begin{equation}
\label{eq:lat_vars}
\begin{aligned}
\Ebb \left[y_{ji} | \x_i \right] &= g_j \left(\a_j^\top \x_i \right) \\
\Rightarrow \Ebb \left[\y_i | \x_i \right] &= \g \left(\A \x_i \right).
\end{aligned}
\end{equation}
For the remainder of this paper, we make an assumption that all $g_j=g$ for notational simplicity, though the same analysis readily extends to the case where $g_j$ are distinct.

Now, let us introduce a set of latent variables $\z_i\in\Rbb^r$ with $r\ll p$ and the corresponding linear parameter $\B=(\b_1\ldots \b_m)^\top \in\Rbb^{m\times r}$ (note we can incorporate a linear offset by augmenting the variable $\z\leftarrow (\z^\top \; 1)^\top$ and adding the linear offset as a column of $\B$). This leads to the asymptotic maximum likelihood estimate,
\begin{equation}
\label{eq:SIM_MLE_full}
\begin{aligned}
\left(\overline{g},\overline{\A},\overline{\B}\right) = \begin{aligned}[t]& \argmin[g,\A,\B] F_2\left(\y_i,\x_i,\z_i,g,\A,\B\right) \\ & \; \textrm{s.t. } g=\nabla G \in\Ccal^{1}, \end{aligned}
\end{aligned}
\end{equation}
where 
{\small\begin{align}
	\label{eq:SIM_MLE_full_obj}
	F_2(\y_i,\x_i,\z_i,g,\A,\B) \overset{\Delta}{=} \Ebb&\left[ \sum\limits_{j=1}^{m}\left[G_*(y_j) + G(\a_j^\top\x_i+\b_j^\top\z_i)\right] \right. \nonumber \\ &\qquad\left.\vphantom{\sum\limits_{j=1}^{m}}- \y^\top (\A \x_i+\B\z_i) \right].
	\end{align}}

Now consider the case in which the true distribution remains the same, but we only observe $\x_i$ and not $\z_i$,
{\small
	\begin{align}
	\label{eq:SIM_MLE_latent}
	(\what{g},\what{\A}) = & \argmin[g,\A] F_3(\y_i,\x_i,g,\A) \overset{\Delta}{=}
	\begin{aligned}[t]&\Ebb\left[ \sum\limits_{j=1}^{m}\left[G_*(y_{ji}) + G(\a_j^\top\x_i)\right] \right. \nonumber \\
	& \; \qquad \quad\left.\vphantom{\sum\limits_{j=1}^{m}}- \y_i^\top (\A \x_i) \right]
	\end{aligned}\\
	& \; \textrm{s.t. } g=\nabla G \in\Ccal^{1}.
	\end{align}
}We now propose a relation between the two models in~\eqref{eq:SIM_MLE_full} and~\eqref{eq:SIM_MLE_latent}, which will finally lead to the SILVar model. Here we present the abridged theorem, and relegate the full expressions and derivation to Appendix~\ref{app:thm}. To establish notation, let primes ( $ '$) denote derivatives, hats ( $\what{ }$ ) denote a parameter estimate with only observed variables, overbars ($\overline{\vphantom{\a}\hphantom{\a}}$) denote an underlying parameter estimate when we have access to both observed and latent variables, and we drop the subscripts from the random variables $\x$ and $\z$ to reduce clutter.
\begin{thm}
	\label{thm:equiv}
	Assume that $\what{g}'(0)\ne 0$ and that $|\what{g}''|\le J$ and $|\overline{g}''|\le J$ for some $J<\infty$. Furthermore, assume in models~\eqref{eq:SIM_MLE_full_obj} and~\eqref{eq:SIM_MLE_latent} that $\max_j \left(\|\what{\a}_j\|_1,\|\overline{\a}_j\|_1 +\|\overline{\b}_j\|_2 \right)\le k,$
	{\small \begin{equation*}
		\max\left( \Ebb[\|\x\|_2 \|\x\|_\infty^2], \Ebb[\|\x\|_2\|\x\|_\infty \|\z\|_2], \Ebb[\|\x\|_2 \|\z\|_2^2] \right) \le s_{Nr},
		\end{equation*}
	}where subscripts in $s_{Nr}$ indicate that the bounds may grow with the values in the subscript.
	Then, the parameters $\what{\A}$ and $\overline{\A}$ from models~\eqref{eq:SIM_MLE_full_obj} and~\eqref{eq:SIM_MLE_latent} are related as
	\begin{equation*}
	\what{\A}= q(\overline{\A}+\L) + \E,
	\end{equation*}
	where $q=\frac{\overline{g}'(0)}{\what{g}'(0)}$, $\boldsymbol{\mu}_\x=\Ebb[\x_i]$,
	\begin{equation*}
	\begin{aligned}
	&\L=\Big(\overline{\B}\Ebb[\z\x^\top] + \left(\overline{\g}(\0)- \what{\g}(\0)\right)\boldsymbol{\mu}_\x^\top  \Big)\left(\Ebb[\x\x^\top]\right)^{\dagger}\\
	\Rightarrow&\textrm{rank}(\L)\le r+1,
	\end{aligned}
	\end{equation*}
	and $\E = \what{\A} - q(\overline{\A}+\L)$ is bounded as
	\begin{equation*}
	\!\begin{aligned}[t] \frac{1}{MN}\|\E\|_F & \le \! \frac{2J \sigma_\ell  \sqrt{N}}{\what{g}'(0)M}s_{Nr}k^2,
	\end{aligned}
	\end{equation*}
	where $\sigma_\ell=\left\|\left(\Ebb\left[\x\x^\top \right]\right)^\dagger \right\|_2$, the largest singular value of the pseudo-inverse of the covariance.
\end{thm}
The proof is given in Appendix~\ref{app:thm}. The assumptions require that the 2nd order Taylor series expansion is accurate around the point $\0$, that the model parameters $\overline{\A}$ and $\overline{\B}$ are bounded, and that the distributions generating the data are not too spread out (beyond $\0$ where the Taylor series expansion is performed). These are all intuitively reasonable and unsurprising assumptions.
Though the theorem poses a hard constraint on $|\what{g}''|$, we hypothesize that this is a rather strong condition that can be weakened to be in line with similar models.

The theorem determines certain scaling regimes under which the sparse plus low-rank approximation remains reasonable. For instance, consider the case where $M\sim N$ and the moments scale as $s_{Nr}\sim \sqrt{N}$, which is reasonable given their form (i.e., very loosely, $\|\x\|_\infty \sim 1$ and $\|\x\|_2 \sim \sqrt{N}$ and small latent power $\|\z\|_2 \sim 1$ relative to $N$). Then, to keep the error of constant order, the power of each row of the matrix would need to stay constant $k\sim 1$. If we see $\A$ as a graph adjacency matrix, then, in rough terms, this can correspond intuitively to the case in which the node in-degrees stay constant so that the local complexity remains the same even as the global complexity increases while the network grows. Again, we hypothesize that this overall bound can be tightened given more assumptions (e.g., on the network topology). Thus, we propose the SILVar model,
\begin{equation}
\label{eq:SILVar}
\what{\y} = \what{\g}\left(\left(\what{\A}+\what{\L}\right) \x \right),
\end{equation}
and learn it using the optimization problem,
\begin{align}
\label{eq:SILVar_opt}
(\what{g},\what{\A},\what{\L}) = & \argmin[g,\A,\L] \what{F}_3(\Y,\X,g,\A+\L) +h_1(\A)+h_2(\L) \nonumber  \\ & \quad \textrm{s.t. } g=\nabla G \in\Ccal^{1},
\end{align}
where 
{\small
	\begin{equation}
	\label{eq:SILVar_opt_obj}
	\begin{aligned}
	\what{F}_3(\Y,\X,g,\A) \!\!=\!\! \frac{1}{n}\!\sum\limits_{i=1}^{n}\!\left[ \sum\limits_{j=1}^{m}\!\left[G_{\!*\!}\left(y_{ji} \!\right) \!+\! G\left(\a_j^\top\!\x_i \!\right)\right] \!\!-\! \y_i^\top \!\left(\!\A \x_i\!\right)\! \right],
	\end{aligned}
	\end{equation}
}the empirical version of $F_3$, and $h_1$ and $h_2$ are regularizers on $\A$ and $\L$ respectively. Two natural choices for $h_2$ would be $h_2(\L)=\lambda_2\|\L\|_*$ the nuclear norm and $h_2(\L)=\mathbb{I}\{\|\L\|_*\le \lambda_2\}$ the indicator of the nuclear norm ball, both relating to the nuclear norm of $\L$, since $\L$ is approximately low rank due to the influence of a relatively small number of latent variables. We may choose different forms for $h_1$ depending on our assumptions about the structure of $\A$. For example, if $\A$ is assumed sparse, we may use $h_1(\A)=\lambda_1\|\textrm{v}(\A)\|_1$, the $\ell_1$ norm applied element-wise to the vectorized $\A$ matrix. These examples are extensions to the ``sparse and low-rank'' models, which have been shown under certain geometric incoherence conditions to be identifiable~\cite{chandrasekaran_latent_2012}. In other words, if the sparse component is not too low-rank, and if the low-rank component is not too sparse, then $\A$ and $\L$ can be recovered uniquely.
\subsection{Connection to Related Problems}
\label{sec:SILVar:rel_probs}
In this section, we show how the SILVar model can be used in various problem settings commonly considered throughout the literature.
\subsubsection{Generalized Robust PCA}
Though we posed our initial problem as a regression problem, if we take our measurement vectors to be $\x_i=\e_i$ the canonical basis vectors (i.e., so that the overall data matrix $\X=\I$), then we arrive at
\begin{equation}
\what{\Y}=\what{\g}(\what{\A}+\what{\L}).
\end{equation}
This is precisely the model behind Generalized Robust PCA~\cite{candes_robust_2011}, but with the twist of estimating the link function as well~\cite{ganti_matrix_2015}. What is worth noting is that although we arrived at our model via a regression problem with latent variables, the model itself is also able to describe a problem that arises from very different assumptions on how the data is generated and structured.

We also note that the SILVar model can be modified to share a space with the Generalized Low-Rank (GLR) Models~\cite{udell_generalized_2016}. The GLR framework is able to describe arbitrary types of data with an appropriate choice of convex link function $g$ determined \emph{a priori}, while the SILVar model is restricted to a certain continuous class of convex link functions but aims to learn this function. The modification is simply a matrix factorization $\L=\U\V$ (and ``infinite'' penalization on $\A$). The explicit factorization makes the problem non-convex but instead block convex, which still allows for alternating convex steps in $\U$ with fixed $\V$ (and vice versa) to tractably reach local minima under certain conditions. Nonetheless, due to the non-convexity, further discussion of this particular extension will be beyond the scope of this paper.
\subsubsection{Extension to Autoregressive Models}
We can apply the SILVar model to learn from time series as well. Consider a set of $N$ time series each of length $K$, $\X\in\Rbb^{N\times K}$. We assume the noise at each time step is independent (note that, with this assumption, the time series are still dependent across time), and take in our previous formation, $\y_i\leftarrow \x_k$ and $\x_i\leftarrow \x_{k-1:k-M}=(\x_{k-1}^\top \ldots \x_{k-M}^\top)^\top$ so that the model of order $M$ takes the form,
\begin{equation}
\label{eq:SILVar_AR}
\what{\x}_k = \g\left(\sum_{i=1}^{M}\left(\A^{(i)}+\L^{(i)}\right) \x_{k-i}\right),
\end{equation}
and learn it using the optimization problem,
{\small
	\begin{equation}
	\label{eq:SILVar_AR_opt}
	(\what{g},\what{\A},\what{\L}) = \begin{aligned}[t]& \argmin[g,\A,\L] \what{F}_4(\X,g,\A+\L) +h_1(\A)+h_2(\L)  \\ & \quad \textrm{s.t. } g=\nabla G \in\Ccal^{1}, \end{aligned}
	\end{equation}
}where $\A=\left(\A^{(1)} \ldots \A^{(M)}\right)$ and $\L=\left(\L^{(1)} \ldots \L^{(M)}\right)$ and
{\small
	\begin{equation*}
	\label{eq:SILVar_AR_opt_obj}
	\begin{aligned}
	\what{F}_4(\X,g,\A) \!=\!\! \frac{1}{K \!-\! M}\!\!\sum\limits_{k \!=\! M \!+\! 1}^{K}\!\!\!\begin{aligned}[t]&\left[ \sum\limits_{j=1}^{N}\left[G_*(x_{jk}) \!+\! G\left(\sum\limits_{i=1}^M \a_j^{(i)}\x_{k-i}\right)\right] \right.\\
	&\qquad\quad \left. \vphantom{\sum\limits_{j=1}^{m}} - \x_k^\top \left(\sum\limits_{i=1}^M \A^{(i)} \x_{k-i} \right) \right],\end{aligned}
	\end{aligned}
	\end{equation*}
}where $\A^{(i)}\!=\!\left(\a_1^{(i)}\ldots\a_N^{(i)}\right)^\top$, similarly to before. Note that the analysis in the previous section follows naturally in this setting, so that here $\textrm{rank}(\L_i)\le r+1$. Then, the matrix $\A$ may be assumed to be group sparse, relating to generalized notions of Granger Causality~\cite{granger_causality_1988,bolstad_causal_2011}, and one possible corresponding choice of regularizer taking the form $h_1(\A)=\lambda_1\sum\limits_{i,j}\left\|\left(a^{(1)}_{ij} \ldots a^{(M)}_{ij}\right)\right\|_2$. 

Another structural assumption could be that of the Causal Graph Process model~\cite{mei_signal_2015-1}, inspired by Signal Processing on Graphs~\cite{sandryhaila_discrete_2013}, in which $\A^{(i)}$ are matrix polynomials in one underlying matrix $\wtil{\A}$. This framework utilizes the nonconvex regularizer $h_1(\A)=\lambda_1\|\textrm{v}(\A^{(1)})\|_1+\lambda_2\sum\limits_{i\ne j}\|\A^{(i)}\A^{(j)}-\A^{(j)}\A^{(i)}\|_F^2$ to encourage both sparsity and commutativity, which is satisfied if $\A^{(i)}$ are all matrix polynomials in the same matrix. Since this particular regularization is again block convex, convex steps can still be taken in each $\A^{(i)}$ with all other blocks $\A^{(j)}$ for $j\ne i$ fixed, for a computationally tractable algorithm to reach a local minimum under certain conditions. However, further detailed discussion will remain outside the scope of this paper.

The hidden variables in this time series model can even possibly model underlying trends in the data. For example, one year of daily temperature measurements across a country could be related through a graph based on geographical and meteorological features, but all exhibit the same significant trend due to the changing seasons. In previous work, a standard pipeline is to detrend the data as a preprocessing step, and then estimate or use a graph to describe the variations of the data on top of the underlying trends~\cite{sandryhaila_discrete_2013,sandryhaila_discrete_2014,mei_signal_2017}. Instead, the time series can also be modeled as a modified autoregressive process, depending on a low-rank trend $\L'=\left(\boldsymbol{\ell}'_1 \; \ldots \; \boldsymbol{\ell}'_K\right)\in\Rbb^{N\times K}$ and the variations of the process about that trend,
\begin{equation}
\begin{aligned}
\label{eq:trend}
&\what{\x}_k=g\left(\boldsymbol{\ell}'_k+\sum_{i=1}^{M}\A^{(i)}\left(\x_{k-i}-\boldsymbol{\ell}'_{k-i}\right)\right).
\end{aligned}	
\end{equation}
Substituting this into Equation~\eqref{eq:SILVar_AR} yields
\begin{equation}
\begin{aligned}
\label{eq:trend2}
&\boldsymbol{\ell}'_k \!+\! \sum_{i=1}^{M}\!\A^{(i)}\!\left(\x_{k \!-\! i} \!-\! \boldsymbol{\ell}'_{k \!-\! i}\right) \!=\! \sum_{i=1}^{M}\left(\!\A^{(i)} \!+\! \L^{(i)}\!\right) \x_{k\!-\!i}\\
\Rightarrow&\sum_{i=1}^{M}\L^{(i)}\x_{k-i} = \boldsymbol{\ell}'_{k}-\sum_{i=1}^{M}\A^{(i)}\boldsymbol{\ell}'_{k-i}
\end{aligned}	
\end{equation}
Thus we estimate the trend using ridge regression for numerical stability purposes but without enforcing $\L'$ to be low rank. We can accomplish this via the simple optimization,
\begin{equation}
\begin{aligned}
\label{eq:trend_opt}
\what\L' \!=\! \argmin[\L'] &\!\sum\limits_{k\!=\!M\!+\!1}^{K} \left\|\boldsymbol{\ell}'_{k} \!-\! \sum_{i=1}^{M}\A^{(i)}\boldsymbol{\ell}'_{k\!-\!i} \!-\! \sum_{i=1}^{M}\L^{(i)}\x_{k\!-\!i} \right\|_2^2\\
& \; +\lambda\|\L'\|_F^2
\end{aligned}	
\end{equation}
with $\lambda$ being the regularization parameter. In this way, the extension of SILVar to autoregressive models can allow joint estimation of the effects of the trend and of these variations supported on the graph, as will be demonstrated via experiments in Section~\ref{sec:exp}.

\section{Efficiently Learning SILVar Models}
\label{sec:learn_SILVar}
In this section, we describe the formulation and corresponding algorithm for learning the SILVar model. Surprisingly, in the single-task setting, learning a SIM is jointly convex in $g$ and $\a$ as demonstrated in~\cite{acharyya_parameter_2015}. The pseudo-likelihood functional $\what{F}_3$ used for learning the SILVar model in~\eqref{eq:SILVar_AR_opt} is thus also jointly convex in $g$, $\A$, and $\L$ by a simple extension from the single-task regression setting. 

\begin{lem}[Corollary of Theorem 2 of~\cite{acharyya_parameter_2015}]
	The $\what{F}_3$ in the SILVar model learning problem~\eqref{eq:SILVar} is jointly convex in $g$, $\A$, and $\L$.
\end{lem}

This convexity is enough to ensure that the learning can converge and be computationally efficient. Before describing the full algorithm, one detail remains: the implementation of the LMR introduced in Section~\ref{sec:bg:LMR}. 

\subsection{Lipschitz Monotonic Regression}		
\label{sec:learn_SILVar:LMR}
To tackle LMR, we first introduce the related simpler problem of monotonic regression, which is solved by a well-known algorithm, the pooled adjacent violators (PAV)~\cite{dykstra_isotonic_1981}. The monotonic regression problem is formulated as
\begin{equation}
\label{eq:mon_reg}
\begin{aligned}
\textrm{PAV}(\y,\x) &\overset{\Delta}{=} \argmin[\g] \sum\limits_{i=1}^{n}\;\left( g(x_i) - y_i \right)^2\\
& \quad \textrm{s.t. } 0 \le g\left(x_{[j+1]}\right)-g\left(x_{[j]}\right).
\end{aligned}
\end{equation}
The PAV algorithm has a complexity of $O(n)$, which is due to a single complete sweep of the vector $\y$. The monotonic regression problem can also be seen as a standard $\ell_2$ projection onto the convex set of monotonic functions.

We introduce a simple generalization to the monotonic regression,
\begin{equation}
\label{eq:gen_mon_reg}
\begin{aligned}
\textrm{GPAV}_\t(\y,\x) &\overset{\Delta}{=} \argmin[\g] \sum\limits_{i=1}^{n}\;\left( g(x_i) - y_i \right)^2\\
& \quad \textrm{s.t. } t_{[j+1]} \le g\left(x_{[j+1]}\right)-g\left(x_{[j]}\right) \\
&\qquad\qquad \textrm{for } j=1,\ldots, n-1.
\end{aligned}
\end{equation}
This modification allows weaker (if $t_i<0$) or stronger (if $t_i>0$) local monotonicity conditions to be placed on the estimated function $g$.
Let $t_{[1]}=0$ and $s_{[i]}=\sum\limits_{j=1}^i t_{[j]}$, and in vectorized form,
\begin{equation}
\label{eq:cusum}
\s\overset{\Delta}{=}\textrm{cusum}(\t).
\end{equation}
Then, we can rewrite,
{\small
	\begin{equation}
	\label{eq:gen_mon_reg2}
	\begin{aligned}
	\textrm{GPAV}_\t(\y,\x) &= \argmin[\g] \sum\limits_{i=1}^{n}\;\left( g(x_i) - s_i - \left(y_i - s_i\right) \right)^2\\
	& \quad \textrm{s.t. } 0 \le g\left(x_{[j+1]}\right) \!-\! s_{[j+1]} \!-\! \left(g\left(x_{[j]}\right) \!-\! s_{[j]}\right) \\
	&\qquad\qquad \textrm{for } j=1,\ldots, n-1,
	\end{aligned}
	\end{equation}
}which leads us to recognize that
\begin{equation}
\label{eq:gpav_pav}
\begin{aligned}
\textrm{GPAV}_\t(\y,\x)= \textrm{PAV}(\y-\s, \x)+ \s.
\end{aligned}
\end{equation}

Returning to LMR~\eqref{eq:Lip_mon_reg}, we pose it as the projection onto the intersection of 2 convex sets, the monotonic functions and the functions with upper bounded differences, for which each projection can be solved efficiently using GPAV (for the second set, we can use GPAV on the negated function and negate the result). Thus to perform this optimization efficiently utilizing PAV as a subroutine, we can use an accelerated Dykstra's Algorithm~\cite{lopez_acceleration_2016}, which is a general method to find the projection onto the non-empty intersection of any number of convex sets. The accelerated algorithm can has a geometric convergence rate for $O(\log n)$ passes, with each pass utilizing PAV with $O(n)$ cost, for a total cost of $O(n\log n)$. We make a brief sidenote that this is better than the $O(n^2)$ in~\cite{yeganova_isotonic_2009} and simpler than the $O(n\log n)$ in~\cite{kakade_efficient_2011} achieved using a complicated tree-like data structure, and has no learning rate parameter to tune compared to an ADMM-based implementation that would also yield $O(n\log n)$. We provide the steps of LMR in Algorithm~\ref{alg:LMR} as a straightforward application of the the accelerated Dykstra's algorithm, but leave the details of the derivation of the acceleration to~\cite{lopez_acceleration_2016}. In addition, the numerical stability parameter is included to handle small denominators, but in practice we did not observe the demoninator falling below $\varepsilon=10^{-9}$ in our simulations before convergence.

\begin{algorithm}
	\caption{Lipschitz monotone regression (LMR)}
	\label{alg:LMR}
	\begin{algorithmic}[1]
		\State Let $t_1=0$ and compute $t_{[i+1]}=x_{[i+1]} - x_{[i]}$, $\s=\textrm{cusum}(\t)$. Set numerical stability parameter $0<\varepsilon<1$, $error=\infty$, and tolerance $0<\delta<1$.
		\State Initialize $\g_{\ell}^{(0)}=\textrm{PAV}(\y,\x)$, $\g_{u}^{(0)}=-\textrm{PAV}(-\y+\s,\x)+\s$,
		$\v_0=\0$, $\v_1=\g_{u}-\g_{\ell}$, $\w=\g_{u}$, $k=0$
		\While{$error \ge \delta$}
		\State $\g_{\ell}^{(k+1)}\leftarrow \textrm{PAV}\left(\g_{u}^{(k)} - \v_0, \x\right)$
		\State $\g_{u}^{(k+1)}\leftarrow - \textrm{PAV}\left(-(\g_{\ell}^{(k+1)}-\v_1)+\s, \x\right)+\s$
		\State $\v_0\leftarrow \g_{\ell}^{(k+1)} - (\g_u^{(k+1)}-\v_0)$
		\State $\v_1\leftarrow \g_{u}^{(k+1)} - (\g_{\ell}^{(k+1)}-\v_1)$
		\If{$\Big| \!\|\g_{\ell}^{(k)} \!\!-\! \g_{u}^{(k)}\|_2^2 \!+\! (\g_\ell^{(k\!+\!1)} \!\!-\! \g_u^{(k\!+\!1)})\!^\top \!(\g_\ell^{(k)} \!\!-\! \g_u^{(k)}) \Big| \!\!\ge\! \varepsilon$}
		\State $\alpha^{(k+1)}\leftarrow \frac{\|\g_{\ell}^{(k)} \!-\! \g_{u}^{(k)}\|_2^2 }{\|\g_{\ell}^{(k)} \!-\! \g_{u}^{(k)}\|_2^2 \!+\! (\g_\ell^{(k+1)} \!-\! \g_u^{(k+1)})\!^\top \!(\g_\ell^{(k)} \!-\! \g_u^{(k)})}$
		\State $\z\leftarrow \g_{u}^{(k)}+\alpha^{(k+1)}(\g_{u}^{(k+1)}-\g_{u}^{(k)})$
		\Else
		\State $\z\leftarrow \frac{1}{2}(\g_{\ell}^{(k+1)}+\g_{u}^{(k+1)})$
		\EndIf 
		\State $error\leftarrow\|\z-\w\|_2$
		\State $\w\leftarrow\z$
		\State $k\leftarrow k+1$
		\EndWhile\\
		\Return $\z$
	\end{algorithmic}
\end{algorithm}
While the GPAV allows us to quickly perform Lipschitz monotonic regression with our particular choice of $\t$, it should be noted that using other choices for $\t$ can easily generalize the problem to bi-H\"{o}lder regression as well, to find functions in the set $\Ccal_{\ell,\alpha}^{u,\beta}=\{g : \forall y>x,\; \ell|y-x|^\alpha \le g(y)-g(x) \le u|y-x|^\beta \}$. This set may prove interesting for further analysis of the estimator and is left as a topic for future investigation.

\subsection{Optimization Algorithms}
\label{sec:learn_SILVar:opt_alg}
With LMR in hand, we can outline the algorithm for solving the convex problem~\eqref{eq:SILVar_opt}. This procedure can be performed for general $h_1$ and $h_2$, but proximal mappings are efficient to compute for many common regularizers. Thus, we describe the basic algorithm using gradient-based proximal methods (e.g., accelerated gradient~\cite{odonoghue_adaptive_2013}, and quasi-Newton~\cite{schmidt_optimizing_2009}), which require the ability to compute a gradient and the proximal mapping.

In addition, we can compute the objective function value for purposes of backtracking~\cite{nocedal_numerical_1999}, evaluating algorithm convergence, or computing error (e.g., for validation or testing purposes). While our optimization outputs an estimate of $g$, the objective function depends on the value of $G$ and its conjugate $G_*$. Though we cannot necessarily determine $G$ or $G_*$ uniquely since $G+c$ and $G$ both yield $g$ as a gradient for any $c\in\Rbb$, the value of $G(x)+G_*(y)$ is unique for a fixed $g$. To see this, consider $\wtil{G}=G+c$ for some constant $c$. Then,
\begin{equation}
\label{eq:G_unique}
\begin{aligned}
\wtil{G}(x)+\wtil{G}_*(y) &=G(x)+c+\max_{z}[zy-\wtil{G}(z)]\\
& = G(x)+c +\max_{z}[zy-G(z)-c]\\
& = G(x) + \max_{z}[zy-G(z)] \\
& = G(x)+G_*(y)
\end{aligned}
\end{equation}

This allows us to compute the objective function by performing the cumulative numerical integral of $\what{\g}$ on points $\textrm{v}(\boldsymbol{\Theta})$ (e.g., \texttt{G=cumtrapz(theta,ghat)} in Matlab). Then, a discrete convex conjugation (also known as the discrete Legendre transform (DLT)~\cite{lucet_faster_1997}) computes $G_*$.

We did not notice significant differences in performance accuracy due to our implementation using quasi-Newton methods and backtracking compared to an accelerated proximal gradient method, possibly because both algorithms were run until convergence. Thus, we show only one set of results. However, we note that the runtime was improved by implementing the quasi-Newton and backtracking method.

\begin{algorithm}
	\caption{Single Index Latent Variable (SILVar) Learning}
	\label{alg:SILVar}
	\begin{algorithmic}[1]
		\State Initialize $\what{\A}=\0$, $\what{\L}=\0$
		\While{not converged} Proximal Methods
		\State Computing gradients:
		\begin{equation*}
		\begin{aligned}
		\boldsymbol{\Theta} &\leftarrow (\what{\A}+\what{\L})\X\\
		\what{\g} &\leftarrow \textrm{LMR}(\textrm{v}(\Y),\textrm{v}(\boldsymbol{\Theta}))\\
		\nabla_{\A} F_3=\nabla_{\L} F_3&=\sum_{i\in\Ical}\left( \what\g(\boldsymbol{\theta}_i)-\y_i \right)\x_i^\top
		\end{aligned}
		\end{equation*}
		\State Optionally compute function value:
		\begin{equation*}
		\begin{aligned}
		\what{\G} &\leftarrow \texttt{cumtrapz}(\textrm{v}(\boldsymbol{\Theta}),\what{\g})\\
		\what{\G}_* &\leftarrow \textrm{DLT}(\textrm{v}(\boldsymbol{\Theta}),\what{\G},\textrm{v}(\Y))\\
		\what{F}_3 &= \sum\limits_{ij} \what{\G}_*(y_{ij})+\what{\G}(\theta_{ij})-y_{ij}\theta_{ij}
		\end{aligned}
		\end{equation*}
		\EndWhile\\
		\Return $(\what{\g},\A,\L)$
	\end{algorithmic}
\end{algorithm}

Algorithm~\ref{alg:SILVar} describes the learning procedure and details the main function and gradient computations while assuming a proximal operator is given. The computation of the gradient and the update vector depends on the particular variation of proximal method utilized; with stochastic gradients, the set $\Ical\subset \{1,\ldots,n\}$ could be pseudorandomly generated at each iteration, while with standard gradients, $\Ical=\{1,\ldots,n\}$. The $\texttt{cumtrapz}$ procedure takes as input the coordinates of points describing the function $\what{\g}$. The DLT procedure takes as its first two inputs the coordinates of points describing the function $\what{\G}$, and as its third input the points at which to evaluate the convex conjugate $\what{\G}_*$. 

Here, an observant reader may notice the subtle point that the function $G_*$ may only be defined inside a finite or semi-infinite interval. This can occur if the function $g$ is bounded below and/or above, so that $G$ has a minimum/maximum slope, and $G_*$ is infinite for any values below that minimum or above that maximum slope of $G$ (to see this, one may refer back to the definition of conjugation~\eqref{eq:conv_conj}). Fortunately, this does not invalidate our method. It is straightforward to see that $\what{g}(x)=x$ is always a feasible solution and that $\what{G}_*(y)=y^2/2$ is defined for all $y$; thus starting from this solution, with appropriately chosen step sizes, gradient-based algorithms will avoid the solutions that make the objective function infinite. Furthermore, even if new data $\y_j$ falls outside the valid domain for the learned $\what{G}_*$ and we incur an ``infinite'' loss using the model, evaluating $\what{\g}\left(\left(\what{\A}+\what{\B}\right)\x_j\right)$ is still well defined. This problem is not unique to SIM's, as assuming a fixed link function $g$ in a GLM can also incur infinite loss if new data does not conform to modeling assumptions implicit to the link function (e.g., the log-likelihood for a negative $\y$ under a non-negative distribution), and making a prediction using the learned GLM is still well-defined. Practically, the loss for SILVar computed using the DLT may be large, but will not be infinite~\cite{lucet_faster_1997}.
\section{Performance}
\label{sec:perf}
We now provide conditions under which the optimization recovers the parameters of the model.

First let us establish a few additional notations needed.
Let $(\wtil g, \wtil\A, \wtil\L)$ be the best Lipschitz monotonic function, sparse matrix, and low-rank matrix that models the true data generation.
Let $\wtil \A$ be sparse on the index set $S$ of size $|S|=s_\A$ and $\wtil\L=\U\Lambda\V^\top$ be the SVD of $\wtil L$, where $\U\in\Rbb^{M \times r_\L}$, $\Lambda\in\Rbb^{r_\L \times r_\L}$, and $\V\in\Rbb^{N \times r_\L}$ where $r_\L$ is the rank of $\wtil \L$. Then let $\A_S$ be equal to $\A$ on $S$ and $0$ on $S^c$ (so that $\wtil\A_S=\wtil\A$ and $\A_S+\A_{S^c}=\A$), and $\L_R=\L-(\I-\U\U^\top)\L(\I-\V\V^\top)$ and $\L_{R^c}=\L-\L_R$ (so that $\wtil\L_R=\wtil\L$ and $\L_{R^c}+\L_R=\L$). Consider the set of approximately $S$-sparse and $R$-low-rank matrices $\mathcal{B}_\gamma(S,R)=\{(\boldsymbol{\Phi},\boldsymbol{\Psi}): \gamma\|\boldsymbol{\Phi}_{S^c}\|_1+\|\boldsymbol{\Psi}_{R^c}\|_* \le 3(\gamma\|\boldsymbol{\Phi}_{S}\|_1+\|\boldsymbol{\Psi}_{R}\|_*)\}$. Intuitively, this is the set of matrices for which the energy of $\boldsymbol{\Phi}$ is on the same sparsity set as $\wtil\A$ and similarly the energy of $\boldsymbol{\Psi}$ is in the same low-rank space as $\wtil\L$.
Finally, let the marginalization of the loss functional w.r.t. $g$ be $\what{m}(\Y,\X,\A)=\underset{g\in\Ccal^1}{\min} \what{F}_3(\Y,\X,g,\A)$, the value of the marginalized function at the true matrix parameters be $\what{\wtil{g}}=\underset{g\in\Ccal^1}{\min} \what{F}_3(\Y,\X,g,\wtil\A+\wtil\L)$, the Hessian of the marginalized functional w.r.t. the matrix parameter be $\wtil{\Ical} = \nabla^2_{\textrm{v}(\A)}\what{m}(\Y,\X,\wtil\A+\wtil\L)$, and denote the quadratic form in shorthand $\|\v\|_{\wtil{\Ical}}=\v^\top \wtil{\Ical}\v$. 

Now, consider the following assumptions:
\begin{enumerate}
	\item\label{as:RSC} There exists some $\alpha>0$ such that
	\begin{align*}
	\|\boldsymbol{\Phi}+\boldsymbol{\Psi}\|_{\wtil{\Ical}}\ge \alpha\|\boldsymbol{\Phi}+\boldsymbol{\Psi}\|_F^2 \quad \forall (\boldsymbol{\Phi},\boldsymbol{\Psi})\in \mathcal{B}_\gamma(S,R).
	\end{align*}
	This is essentially a Restricted Strong Convexity condition standard for structured recovery~\cite{negahban_unified_2012}.
	\item\label{as:Dev} Let $\what{\wtil{\boldsymbol{\Gamma}}}\in \Rbb^{M\times K}$ be the matrix with columns given by 
	\begin{align*}
	\what{\wtil{\boldsymbol{\Gamma}}}_k= \what{\wtil{g}}((\wtil\A+\wtil\L)\x_k)-\y_k, \quad k=1,\ldots,K.
	\end{align*}
	Then for some $\lambda>0$ and $\gamma>0$,
	\begin{align*}
	&\frac{1}{K}\Big\| \what{\wtil{\boldsymbol{\Gamma}}}\X^\top \Big\|_{\infty} \le \lambda\gamma/2 \qquad \frac{1}{K}\Big\| \what{\wtil{\boldsymbol{\Gamma}}}\X^\top \Big\|_2 \le \lambda/2
	\end{align*}
	where $\|\A\|_\infty$ is the largest element of $\A$ in magnitude, and $\|\A\|_2$ is the largest singular value of $\A$.
	
	This states that the error is not too powerful and is not too correlated with the regression variables, and it is also similar to standard conditions for structured recovery. 
	\item\label{as:Inc} Let $\tau=\gamma\sqrt{\frac{s_\A}{r_\L}}$ and $\mu=\frac{1}{32 r_\L (1+\tau^2)}$, then
	\begin{align*}
	\frac{\max( \|\boldsymbol{\Phi}\|_{2},\|\boldsymbol{\Psi}\|_{\infty}/\gamma) }{\gamma\|\boldsymbol{\Phi}\|_{1}+\|\boldsymbol{\Psi}\|_{*}} \le \mu\quad \forall (\boldsymbol{\Phi},\boldsymbol{\Psi})\in \mathcal{B}_\gamma(S,R).
	\end{align*}
	This is a condition on the incoherence between the sparsity set $S$ and low rank subspace $R$ (see~\cite{chandrasekaran_rank-sparsity_2011,chandrasekaran_latent_2012,candes_matrix_2010,candes_robust_2011} for other similar variations on rank-sparsity incoherence considered previously). Basically, this states that $S$ should be such that $S$-sparse matrices are not approximately low-rank, while $R$ should be such that $R$-low-rank matrices are not approximately sparse.
\end{enumerate}
Then we have the following result,
\begin{thm}
	\label{thm:perf}
	Under Assumptions~\ref{as:RSC}-\ref{as:Inc}, the solution to the optimization problem~\eqref{eq:SILVar_AR_opt} satisfies
	\begin{align}
	&\|\what\A-\wtil\A\|_F \le \frac{3\lambda\gamma\sqrt{s_\A}}{\alpha}\left(2 \!+\! \sqrt{\frac{2}{2 \!+\! \tau^2}}\right)\le\frac{9\lambda\gamma\sqrt{s_\A}}{\alpha}\\
	&\|\what\L-\wtil\L\|_F \le \frac{3\lambda\sqrt{r_\L}}{\alpha}\left(2 \!+\! \sqrt{\frac{2\tau^2}{1 \!+\! 2\tau^2}}\right)\le\frac{9\lambda\sqrt{r_\L}}{\alpha}.
	\end{align}
\end{thm}
We give the proof for this theorem in the Appendix. The theorem relates the performance of the optimization to conditions on the problem parameters. Of course, in practice these conditions are difficult to check, since they require knowledge of ground truth. Even given ground truth, verifying Assumption 3 requires solving another separate non-convex optimization problem. Thus, future directions for work could include investigation into which kinds of sparse matrix (or network) models for $\A$ and low-rank models $\L$ produce favorable scaling regimes in the problem parameters with high probability.

\section{Experiments}
\label{sec:exp}
We study the performance of the algorithm via simulations on synthetically generated data as well as real data. In these experiments, we show the different regression settings, as discussed in Section~\ref{sec:SILVar:rel_probs}, under which the SILVar model can be applied.

\subsection{Synthetic Data}
The synthetic data corresponds to a multi-task regression setting. The data was generated by first creating random sparse matrix $\A_f=(\A \; \A_h)$ where $\B\in\Rbb^{p \times H}$ and $H=\lfloor hp \rfloor$ the number of hidden variables, and $h$ is the proportion of hidden variables. The elements of $\A_f$ were first generated as i.i.d. Normal variables, and then a sparse mask was applied to $\A$ choose $\lfloor m\log_{10}(p) \rfloor$ non-zeros, and $\B$ was scaled by a factor of $\frac{1}{3\sqrt{H}}$. Next, the data $\X_f =(\X^\top \; \Z^\top)^\top$ were generated with each vector drawn i.i.d. from a multivariate Gaussian with mean $\0$ and covariance matrix $\boldsymbol{\Sigma}_f\in\Rbb^{(p+H) \times (p+H)}$. This was achieved by creating $\boldsymbol{\Sigma}_f^{1/2}$ by thresholding a matrix with diagonal entries of $1$ and off-diagonal entries drawn from an i.i.d. uniform distribution in the interval $[-0.5,0.5]$ to be greater than 0.35 in magnitude. Then $\boldsymbol{\Sigma}_f^{1/2}$ was pre-multiplied with a matrix of i.i.d. Normal variables. Finally, we generated $\Y=g(\A_f\X_f)+\W$ for 2 different link functions $g_1(x)=\log(1+e^{c_1 x})$ and $g_2(x)=\frac{2}{1+e^{-c_2 x}}-1$, and $\W$ is added i.i.d. Gaussian noise. Then, the task is to learn $(\what{g},\what\A,\what\L)$ the SILVar model~\eqref{eq:SILVar} from $(\Y,\X)$ without access to $\Z$. As comparison, we use an Oracle GLM model, in which the true $g$ is given but the parameters $(\what\A_o,\what\L_o)$ still need to be learned. We note that while there is a slight mismatch between the true and assumed noise distributions, the task is nonetheless difficult, yet our estimation using the SILVar model can still exhibit good performance with respect to the Oracle.

\begin{figure*}[t]
	\begin{subfigure}[t!]{0.95\columnwidth}
		\includegraphics[width=\columnwidth]{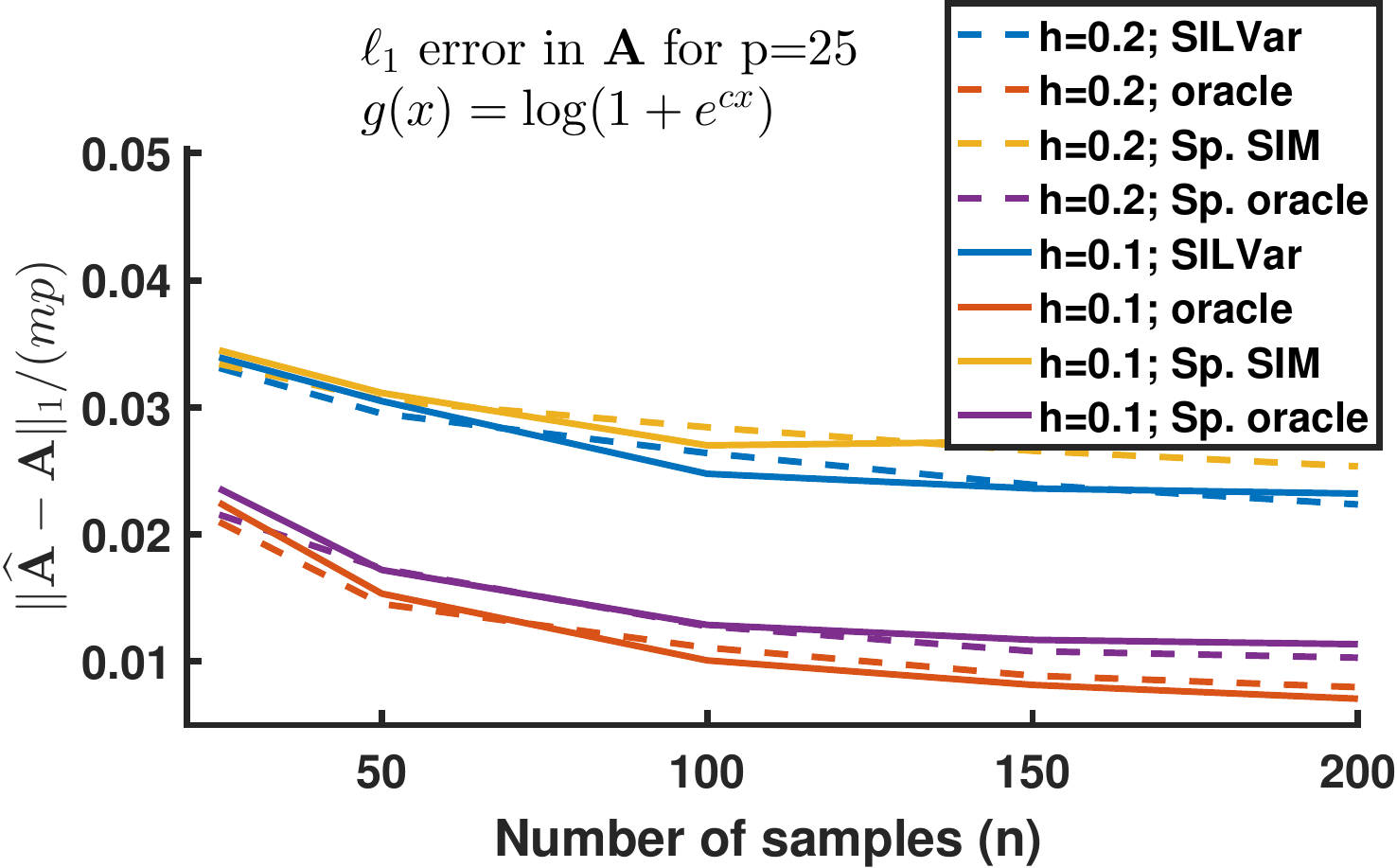}
		\caption{}
		\label{fig:toy:l1_A1_p25}
	\end{subfigure}%
	~
	\begin{subfigure}[t!]{0.95\columnwidth}
		\includegraphics[width=\columnwidth]{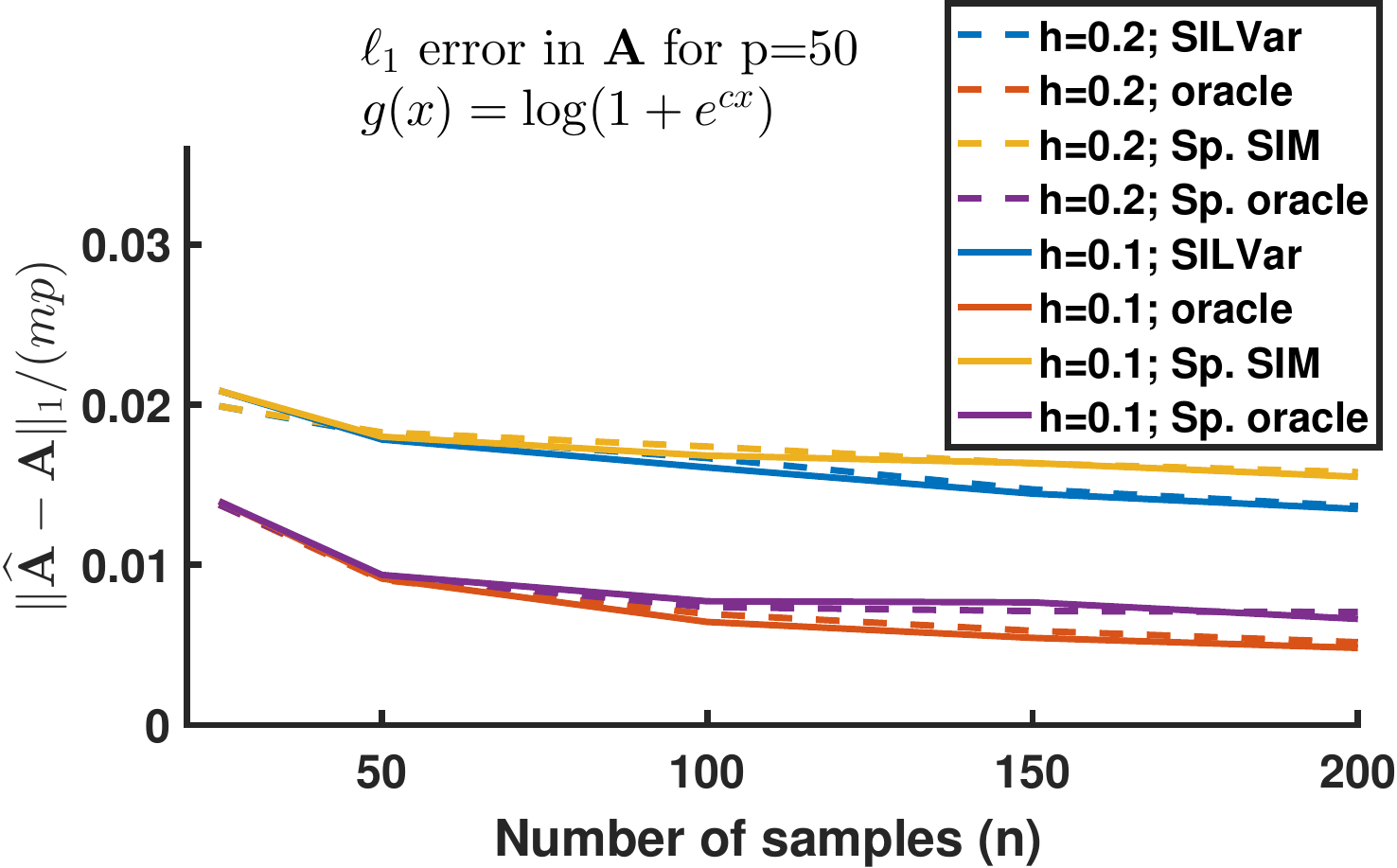}
		\caption{}
		\label{fig:toy:l1_A1_p50}
	\end{subfigure}
	
	\begin{subfigure}[t]{0.95\columnwidth}
		\includegraphics[width=\columnwidth]{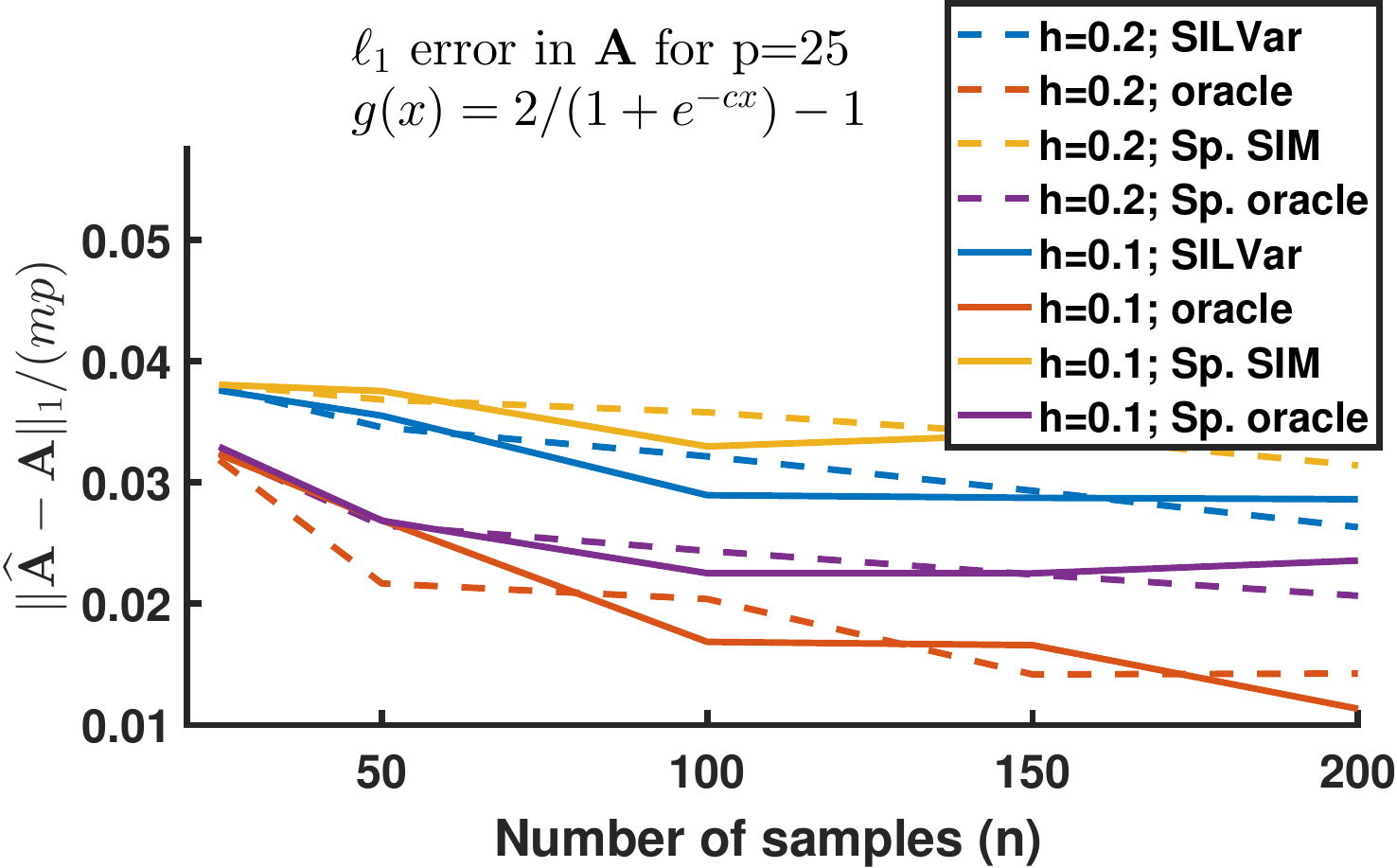}
		\caption{}
		\label{fig:toy:l1_A2_p25}
	\end{subfigure}%
	~
	\begin{subfigure}[t]{0.95\columnwidth}
		\includegraphics[width=\columnwidth]{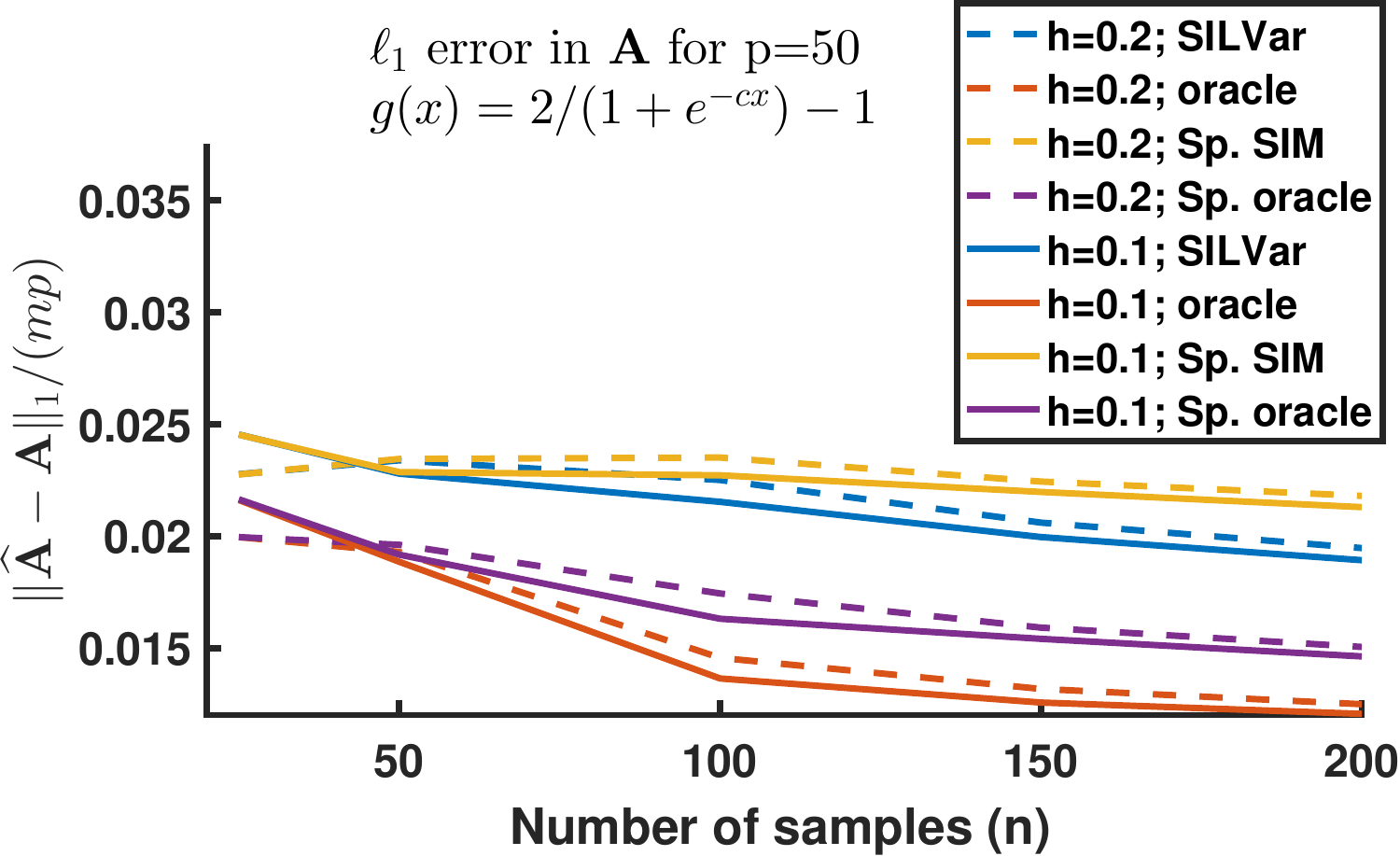}
		\caption{}
		\label{fig:toy:mse_A2}
	\end{subfigure}	
	
	\caption{Different errors for estimating $\A$ in Toy Data}
	\label{fig:toy}		
\end{figure*}

The experiments were carried out across a range of problem sizes. The dimension of $\y_i$ was fixed at $m=25$. The dimension of the observed $\x_i$ was set to $p\in\{25,50\}$, and the proportion of the dimension of hidden $\z_i$ was set to $h\in\{0.1,0.2\}$. The number of data samples was varied in $k\in\{25, 50, 100, 150, 200\}$. Validation was performed using a set of samples generated independently but using the same $\A_f$ and $g$, and using a grid of $(\lambda_S,\lambda_L)\in\left\{10^{i/4} \big| i\in \{-8, -7, \ldots, 7, 8\} \right\}^2$. The whole process of data generation and model learning is repeated 20 times for each experimental condition. The average $\ell_1$ errors between the true $\A$ and the best estimated $\what\A$ (determined via validation) are shown in Figure~\ref{fig:toy}. The first row is for the function $g_1$ and the second row is for the function $g_2$ for the various experimental conditions. The empirical results show that the SILVar model and learning algorithm lose out slightly on performance for estimating the link function $g$ but overall still match the Oracle fairly well in most cases. We also see that SILVar performs better than just the sparse SIM~\cite{ganti_learning_2015}, but that the purely sparse Oracle GLM can still provide an improvement over SILVar for not knowing the link function. We also see several other intuitive behaviors for both the SILVar and Oracle models: as we increase the amount of data $n$, the performance increases. Note that due to the scaling of weights $\B$ by the factor of $1/\sqrt{H}$ to keep the average power $\|\B\|_F$ roughly constant, increasing the proportion of hidden variables (from $h=0.1$ to $h=0.2$) did not directly lead to significant performance decreases. We also note that $g_2$ seems to be somehow more difficult to estimate than $g_1$, and additionally that the performance of the SILVar model w.r.t. the Oracle model is worse with $g_2$ than with $g_1$. This could have something to do with the 2 saturations in $g_2$ as opposed to the 1 saturation in $g_1$, corresponding in an intuitive sense to a more non-linear setting that contains more information needing to be captured while estimating $\what{g}$.

\subsection{Temperature Data}
In this setting, we wish to learn the graph capturing relations between the weather patterns at different cities. The data is a real world multivariate time series consisting of daily temperature measurements (in $^\circ$F) for $365$ consecutive days during the year $2011$ taken at each of $150$ different cities across the continental USA.

Previously, the analysis on this dataset has been performed by first fitting with a $4$th order polynomial and then estimating a sparse graph from an autoregressive model using a known link function $g(x)=x$ assuming Gaussian noise~\cite{mei_signal_2017}. 

Here, we fit the time series using a $2$nd order AR SILVar model~\eqref{eq:SILVar_AR} with regularizers for group sparsity $h_1(\A)=\lambda_1\sum\limits_{i,j}\left\|\left(a^{(1)}_{ij} \ldots a^{(M)}_{ij}\right)\right\|_2$ where $a_{ij}^{(m)}$ is the $ij$ entry of matrix $\A^{(m)}$, and nuclear norm $h_2(\L)=\lambda_2\sum\limits_{i=1}^{M}\left\|\L^{(i)}\right\|_*$. We also estimate the underlying trend using~\eqref{eq:trend_opt}.	

\begin{figure}[t]
	\begin{subfigure}[t!]{0.45\columnwidth}
		\includegraphics[width=\columnwidth]{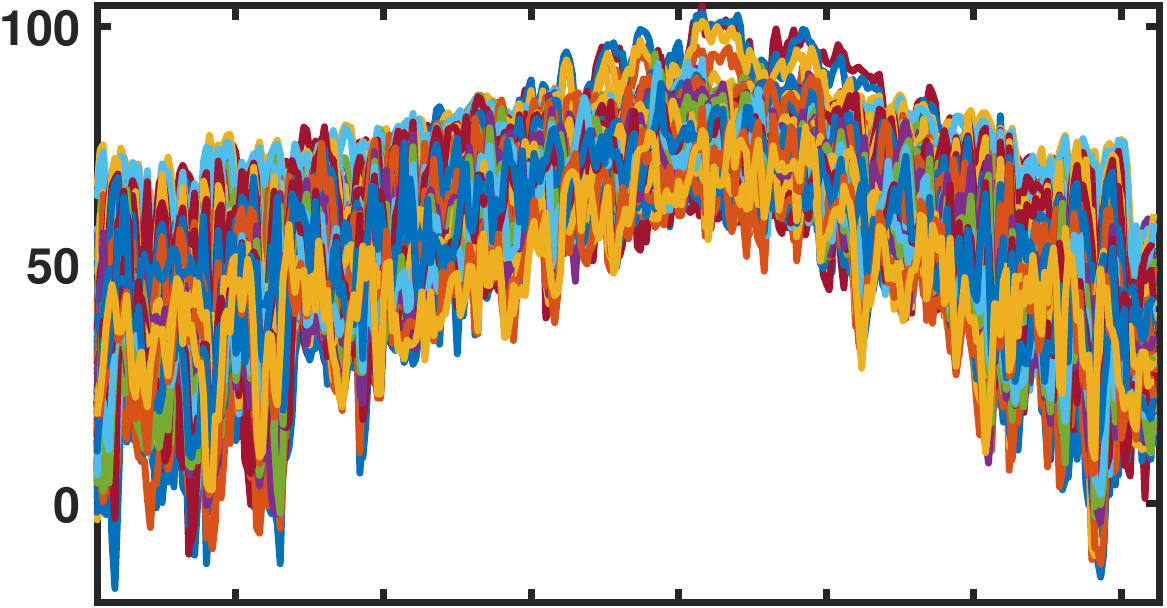}
		\caption{Temperature time series}
		\label{fig:weather_trends:ts}
	\end{subfigure}%
	\begin{subfigure}[t!]{0.45\columnwidth}
		\includegraphics[width=\columnwidth]{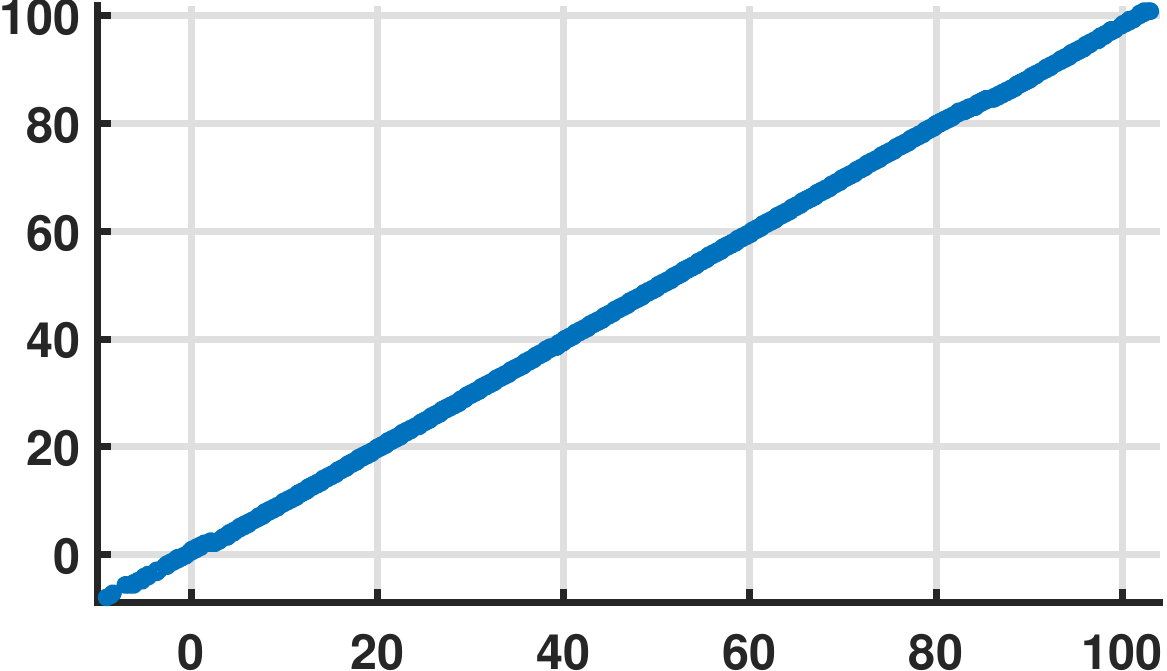}
		\caption{Learned $\what{g}$}
		\label{fig:weather_trends:weather_g_hat}
	\end{subfigure}
	
	
	\begin{subfigure}[t]{0.45\columnwidth}
		\includegraphics[width=\columnwidth]{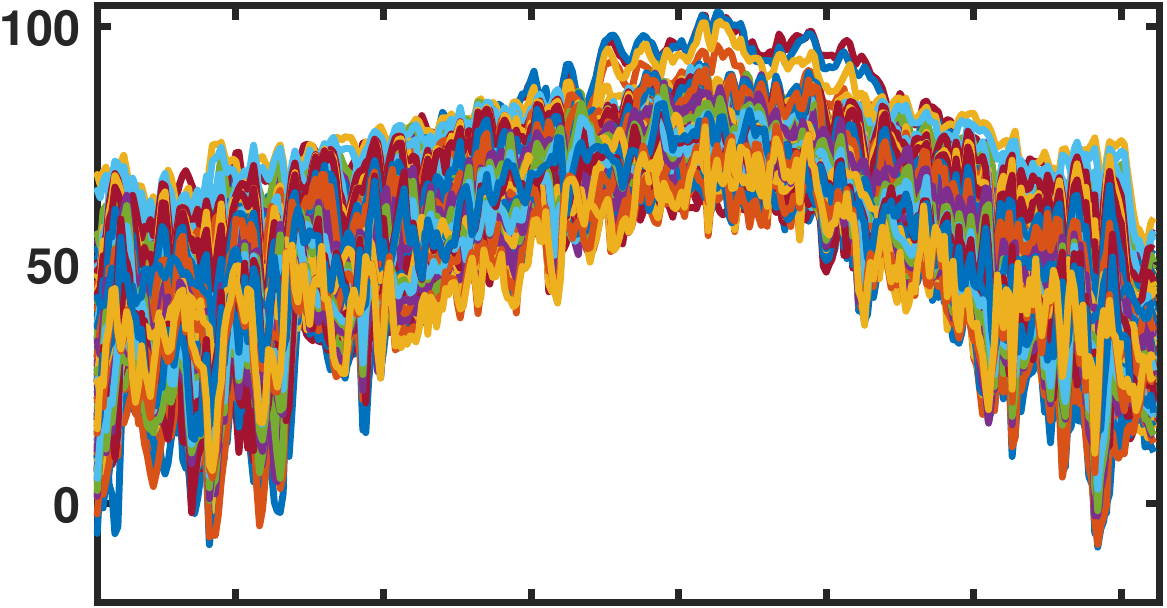}
		\caption{One step prediction}
		\label{fig:weather_trends:full_pred}
	\end{subfigure}%
	~
	\begin{subfigure}[t]{0.45\columnwidth}
		\includegraphics[width=\columnwidth]{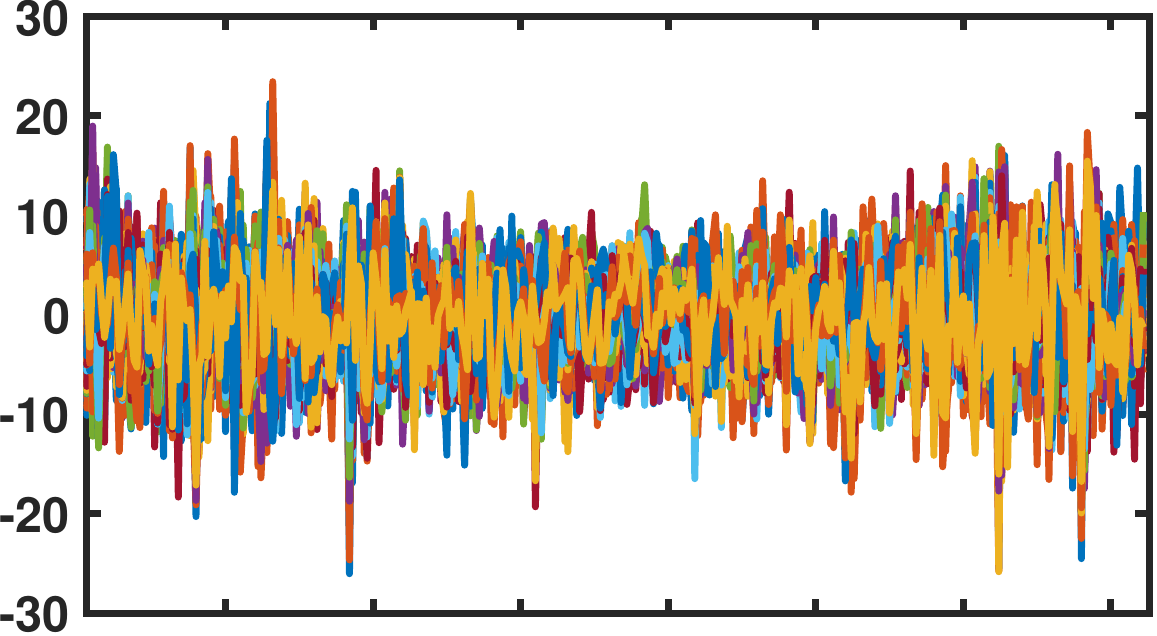}
		\caption{Prediction error}
		\label{fig:weather_trends:full_resid}
	\end{subfigure}	
	
	\begin{subfigure}[t]{0.45\columnwidth}
		\includegraphics[width=\columnwidth]{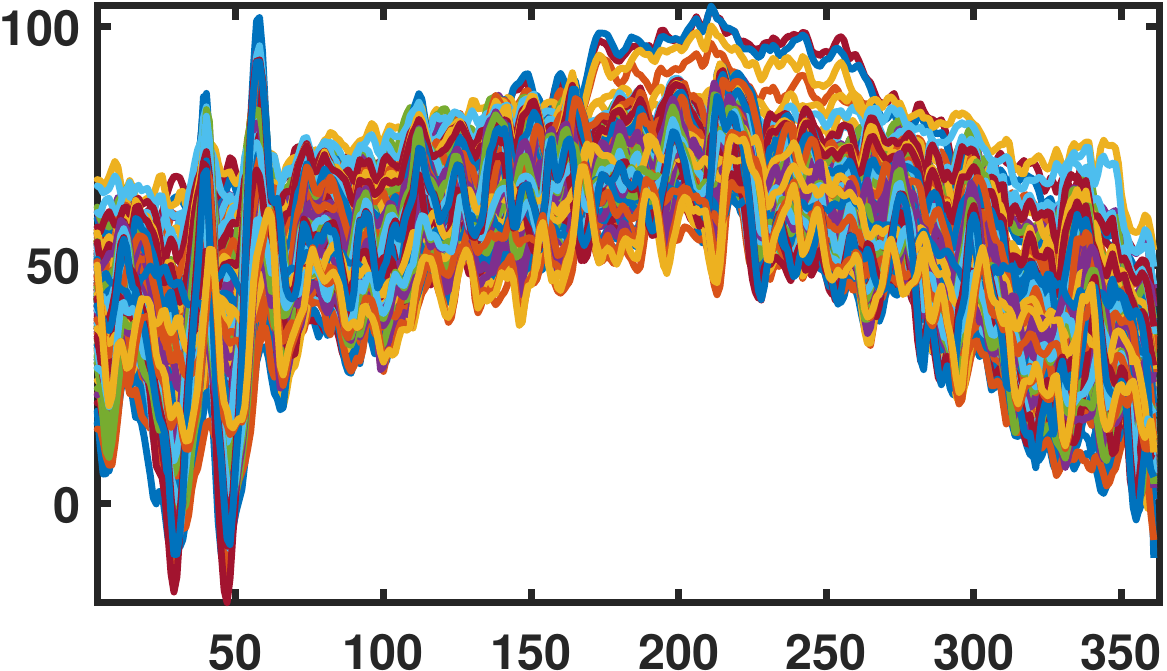}
		\caption{Estimated trend}
		\label{fig:weather_trends:tr_pred}
	\end{subfigure}%
	~
	\begin{subfigure}[t]{0.45\columnwidth}
		\includegraphics[width=\columnwidth]{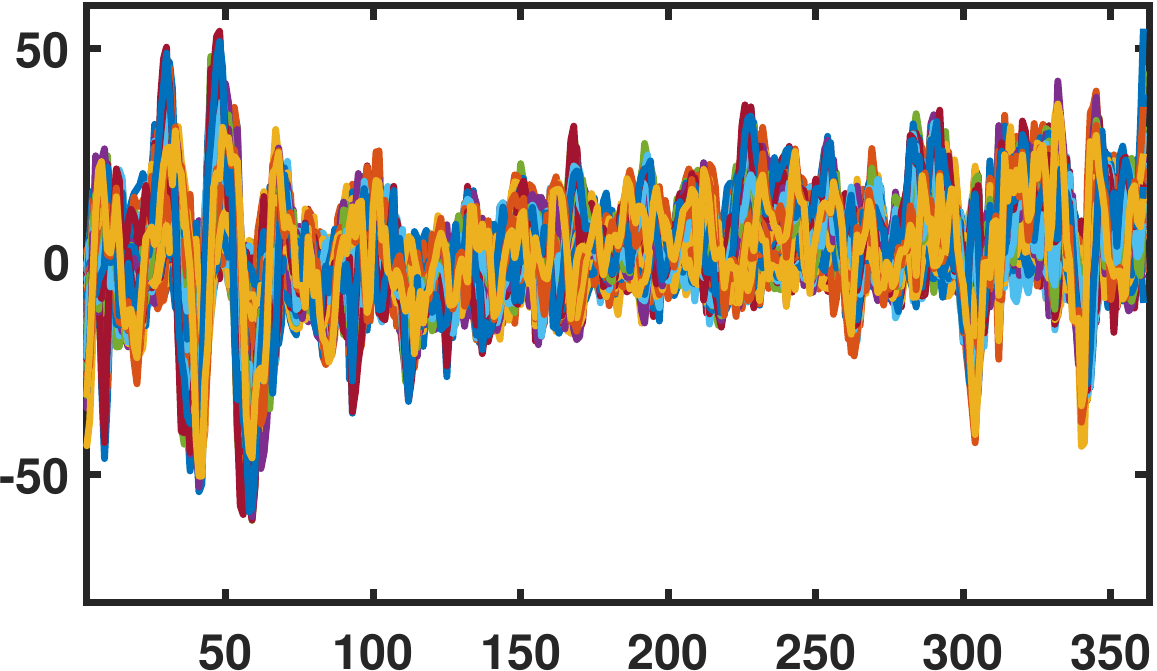}
		\caption{De-trended time series}
		\label{fig:weather_trends:tr_resid}
	\end{subfigure}
	\caption{Time series, link function, trends, and prediction errors computed using the learned SILVar model}
	\label{fig:weather_trends}		
\end{figure}

In Figure~\ref{fig:weather_trends}, we plot the original time series, estimated trends, the estimated network, and residuals. The plots are all from time steps $3$ to $363$, since those are the indices for the trends that we can reliably estimate using the simple procedure~\eqref{eq:trend_opt}. In Figure~\ref{fig:weather_trends:ts}, we show the original time series, which clearly exhibit a seasonal trend winter--summer--winter. Figure~\ref{fig:weather_trends:weather_g_hat} shows the estimated link function $\what{g}$, which turns out to be linear, though a priori we might not intuitively expect a process as complicated as weather to behave this way when sampled so sparsely.
Figures~\ref{fig:weather_trends:full_pred} and~\ref{fig:weather_trends:full_resid} show the full prediction $\what{\x}_k=\sum\limits_{i=1}^M(\what{\A}^{(i)}+\what{\L}^{(i)})\x_{k-i}$ and the residuals $\x_k-\what{\x}_k$, respectively. Note the much lower vertical scale in Figure~\ref{fig:weather_trends:full_resid}.  Figures~\ref{fig:weather_trends:tr_pred} and~\ref{fig:weather_trends:tr_resid} show the trend $\what{\L}'$ learned using~\eqref{eq:trend_opt} and the time series with the estimated trend removed, respectively. The trend estimation procedure captures the basic shape of the seasonal effects on the temperature. Several of the faster fluctuations in the beginning of the year are captured as well, suggesting that they were caused by some larger scale phenomena. Indeed there were several notable storm systems that affected the entire USA in the beginning of the year in short succession~\cite{hedge_summary_2011, hamrick_mid-atlantic_2011}.

\begin{figure}[t!]
	\begin{subfigure}[t]{0.95\columnwidth}
		\includegraphics[width=\columnwidth]{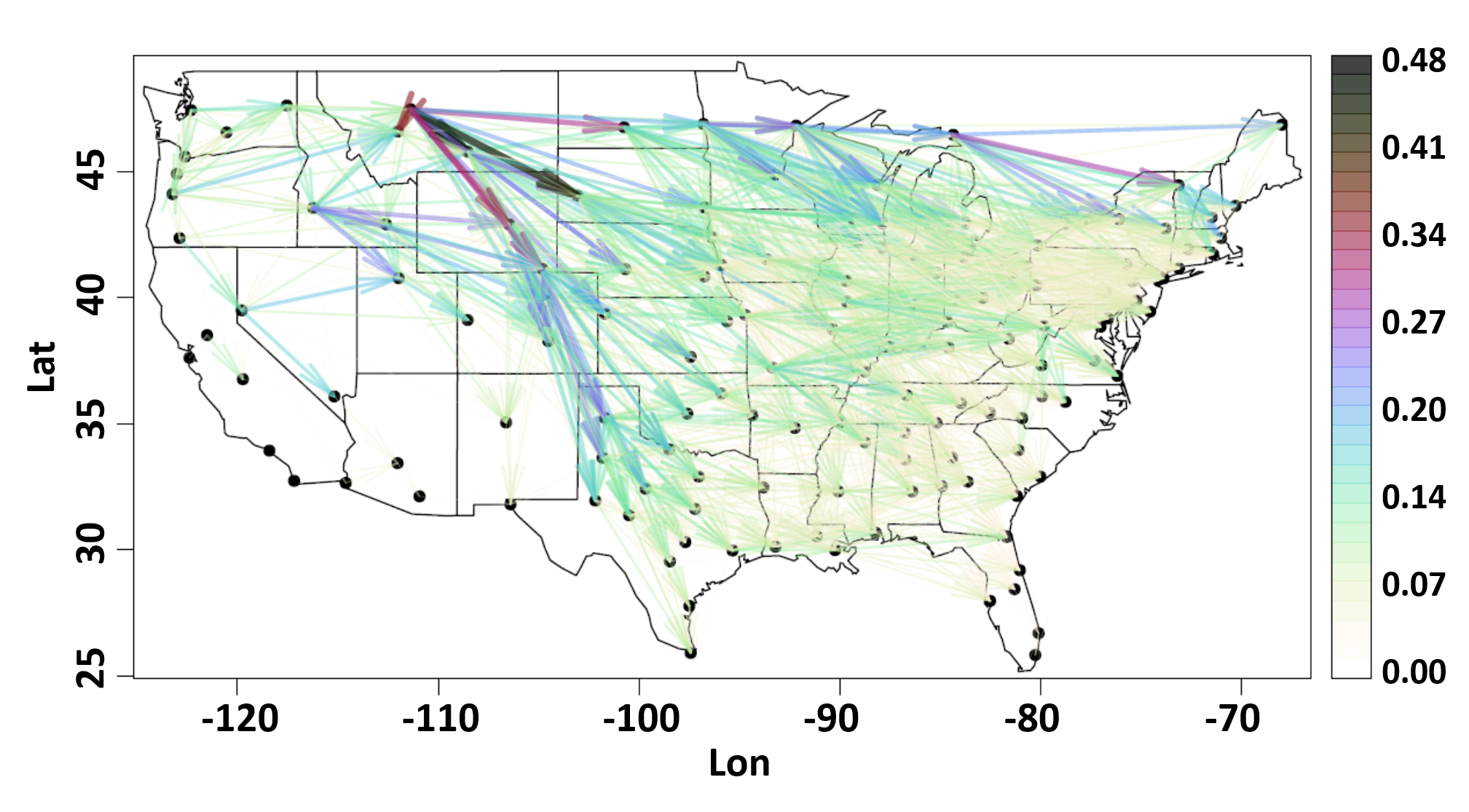}
		\caption{Weather graph learned using SILVar}
		\label{fig:weather_graphs:silvar}
	\end{subfigure}
	\begin{subfigure}[t]{0.95\columnwidth}
		\includegraphics[width=\columnwidth]{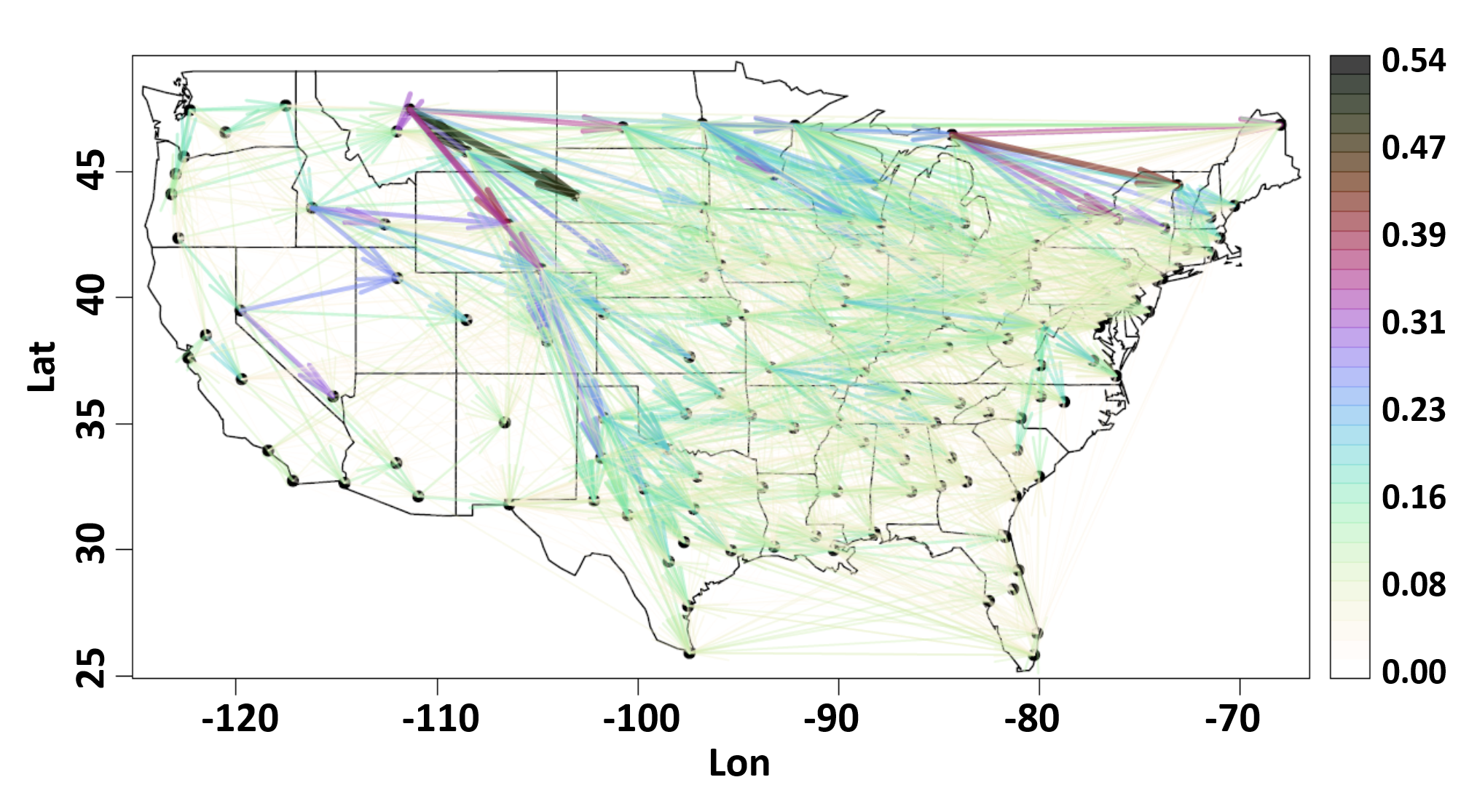}
		\caption{Weather graph learned using Sp. SIM (without low-rank)}
		\label{fig:weather_graphs:nolat}
	\end{subfigure}
	\caption{Learned weather stations graphs}
	\label{fig:weather_graphs}		
\end{figure}

Figure~\ref{fig:weather_graphs} compares two networks $\what{\A}'$ estimated using SILVar and using just sparse SIM without accounting for the low-rank trends, both with the same sparsity level of $12\%$ non-zeros for display purposes, and where $\what{a}'_{ij}=\left\|\left(\what{a}^{(1)}_{ij} \ldots \what{a}^{(M)}_{ij}\right)\right\|_2$.
Figure~\ref{fig:weather_graphs:silvar} shows the network $\what{\A}'$ that is estimated using SILVar. The connections imply predictive dependencies between the temperatures in cities connected by the graph. It is intuitively pleasing that the patterns discovered match well previously established results based on first de-trending the data and then separately estimating a network~\cite{mei_signal_2017}. That is, we see the effect of the Rocky Mountain chain around $-110^\circ$ to $-105^\circ$ longitude and the overall west-to-east direction of the weather patterns, matching the prevailing winds.
In contrast to that of SILVar, the graph estimated by the sparse SIM shown in Figure~\ref{fig:weather_graphs:nolat} on the other hand has many additional connections with no basis in actual weather patterns. Two particularly unsatisfying cities are: sunny Los Angeles, California at $(-118, 34)$, with its multiple connections to snowy northern cities including Fargo, North Dakota at $(-97, 47)$; and Caribou, Maine at $(-68, 47)$, with its multiple connections going far westward against prevailing winds including to Helena, Montana at $(-112, 47)$. These do not show in the graph estimated by SILVar and shown in Figure~\ref{fig:weather_graphs:silvar}.

%
%

\subsection{Bike Traffic Data}
The bike traffic data was obtained from HealthyRide Pittsburgh~\cite{noauthor_healthy_2016}. The dataset contains the timestamps and station locations of departure and arrival (among other information) for each of 127,559 trips taken between 50 stations within the city from May 31, 2015 to September 30, 2016, a total of 489 days. 

We consider the task of using the total number of rides departing from and arriving in each location at 6:00AM-11:00AM to predict the number of rides departing from each location during the peak period of 11:00AM-2:00PM for each day. This corresponds to $\Y\in\mathbb{N}_0^{50 \times 489}$ and $\X\in\mathbb{N}_0^{100 \times 489}$, where $\mathbb{N}_0$ is the set of non-negative integers, and $\A,\L\in\Rbb^{50\times 100}$.
We estimate the SILVar model~\eqref{eq:SILVar} and compare its performance against a sparse plus low-rank GLM model with an underlying Poisson distribution and fixed link function $g_{\scriptsize{\textrm{GLM}}}(x)=\log(1+e^x)$. We use $n\in\{60,120,240,360\}$ training samples and compute errors on validation and test sets of size $48$ each, and learn the model on a grid of $(\lambda_S,\lambda_L)\in\left\{10^{i/4} \big| i\in \{-8, -7, \ldots, 11, 12\} \right\}^2$. We repeat this 10 times for each setting, using an independent set of training samples each time. We compute testing errors in these cases for the optimal $(\lambda_S,\lambda_L)$ with lowest validation errors for both SILVar and GLM models.

We also demonstrate that the low-rank component of the estimated SILVar model captures something intrinsic to the data. Naturally, we expect people's behavior and thus traffic to be different on business days and on non-business days. A standard pre-processing step would be to segment the data along this line and learn two different models. However, as we use the full dataset to learn one single model, we hypothesize that the learned low-rank $\what{\L}$ captures some aspects of this underlying behavior.

\begin{figure}[t]	
	\begin{subfigure}[t]{0.45\columnwidth}
		\includegraphics[width=\columnwidth]{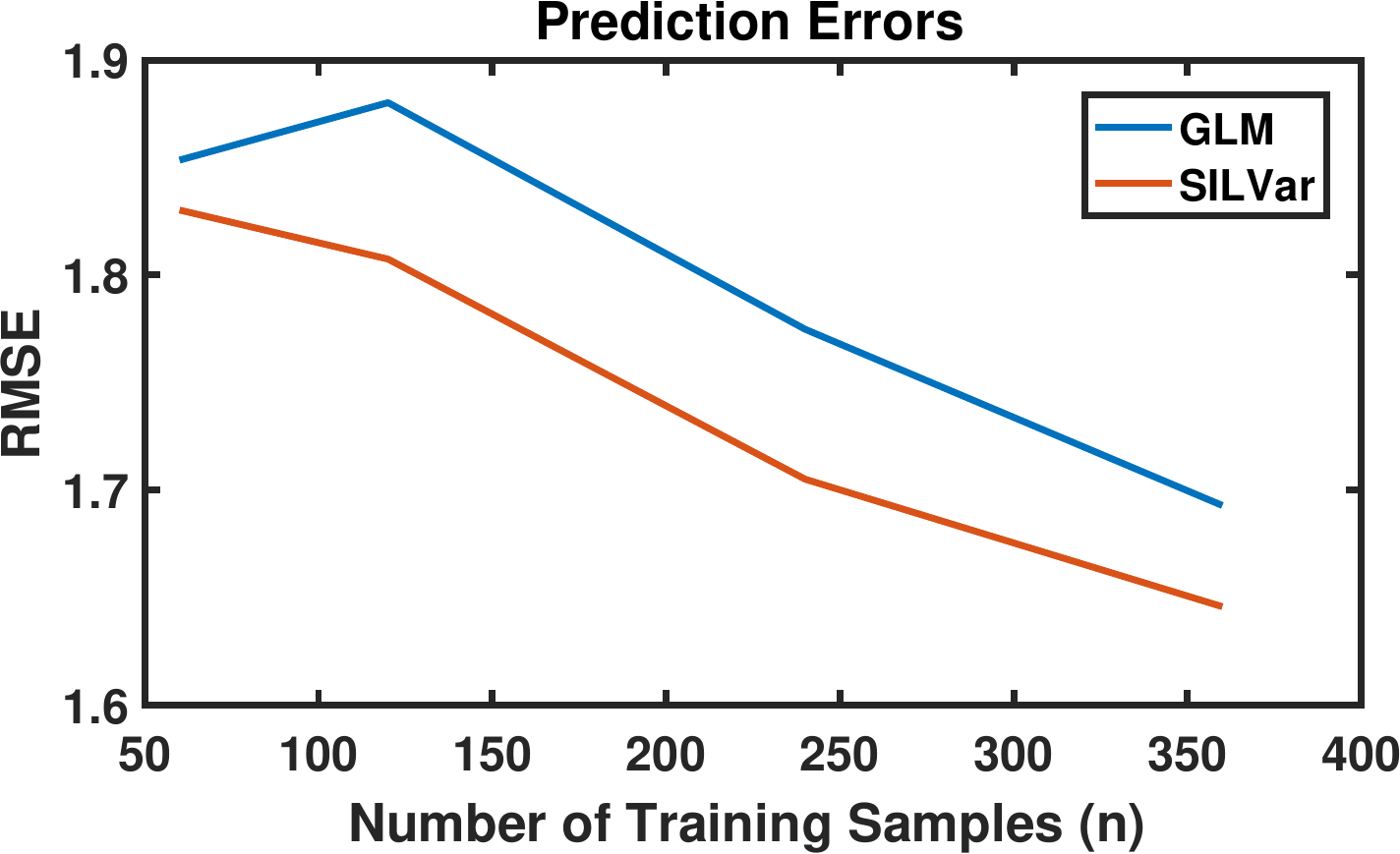}
		\caption{}
		\label{fig:bikes_MSE_link:MSE}
	\end{subfigure}%
	~
	\begin{subfigure}[t]{0.45\columnwidth}
		\includegraphics[width=\columnwidth]{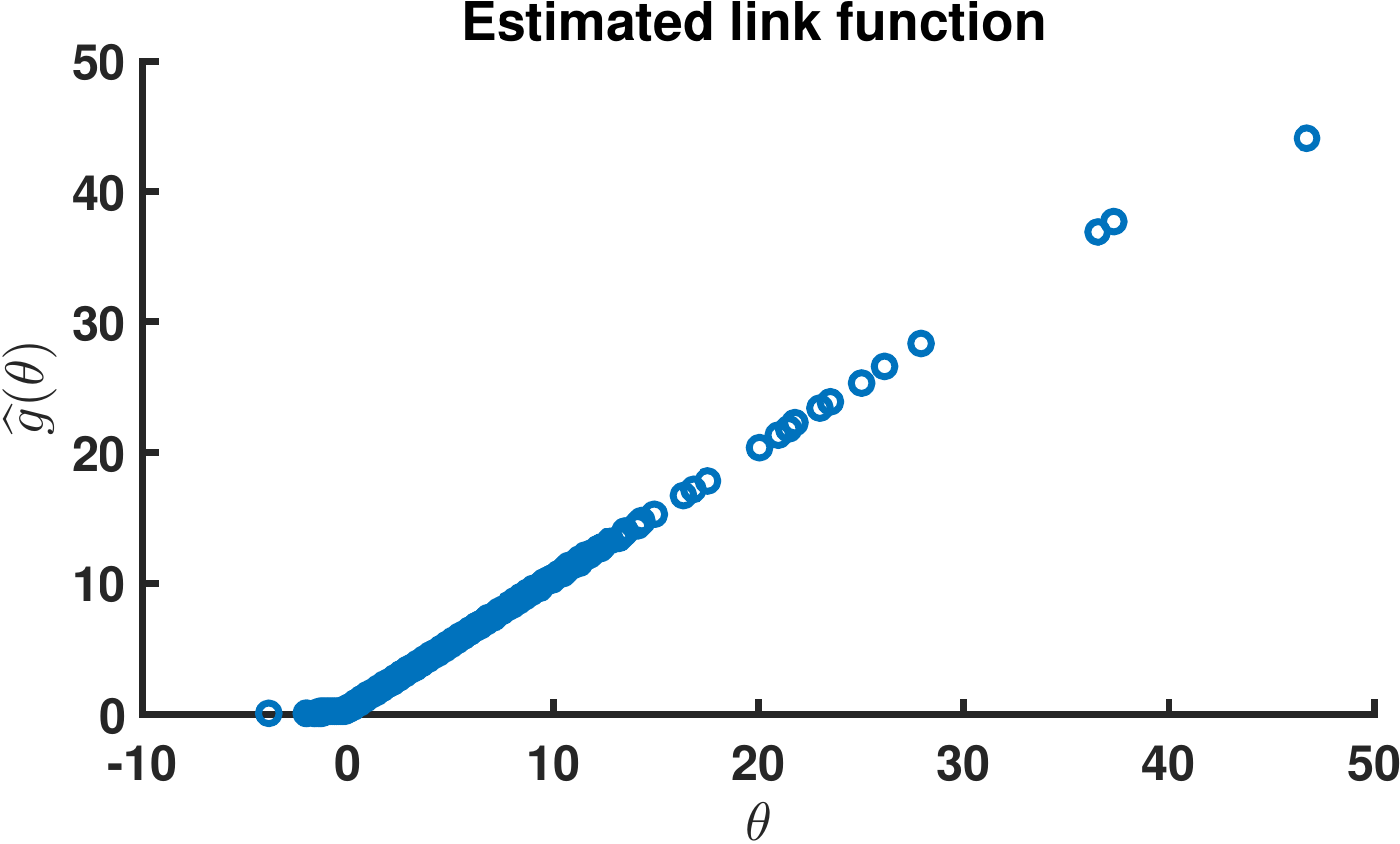}
		\caption{}
		\label{fig:bikes_MSE_link:link}
	\end{subfigure}
	\caption{(a) Root mean squared errors (RMSEs) from SILVar and Oracle models; (b) Link function learned using SILVar model}
	\label{fig:bikes_MSE_link}		
\end{figure}

Figure~\ref{fig:bikes_MSE_link:MSE} shows the test Root Mean Squared Errors (RMSEs) for both SILVar and GLM models for varying training sample sizes, averaged across the 10 trials. We see that the SILVar model outperforms the GLM model by learning the link function in addition to the sparse and low-rank regression matrices. Figure~\ref{fig:bikes_MSE_link:link} shows an example of the link function learned by the SILVar model with $n=360$ training samples. Note that the learned SILVar link function performs non-negative clipping of the output, which is consistent with the count-valued nature of the data. 

\begin{figure}[t]	
	\includegraphics[width=\columnwidth]{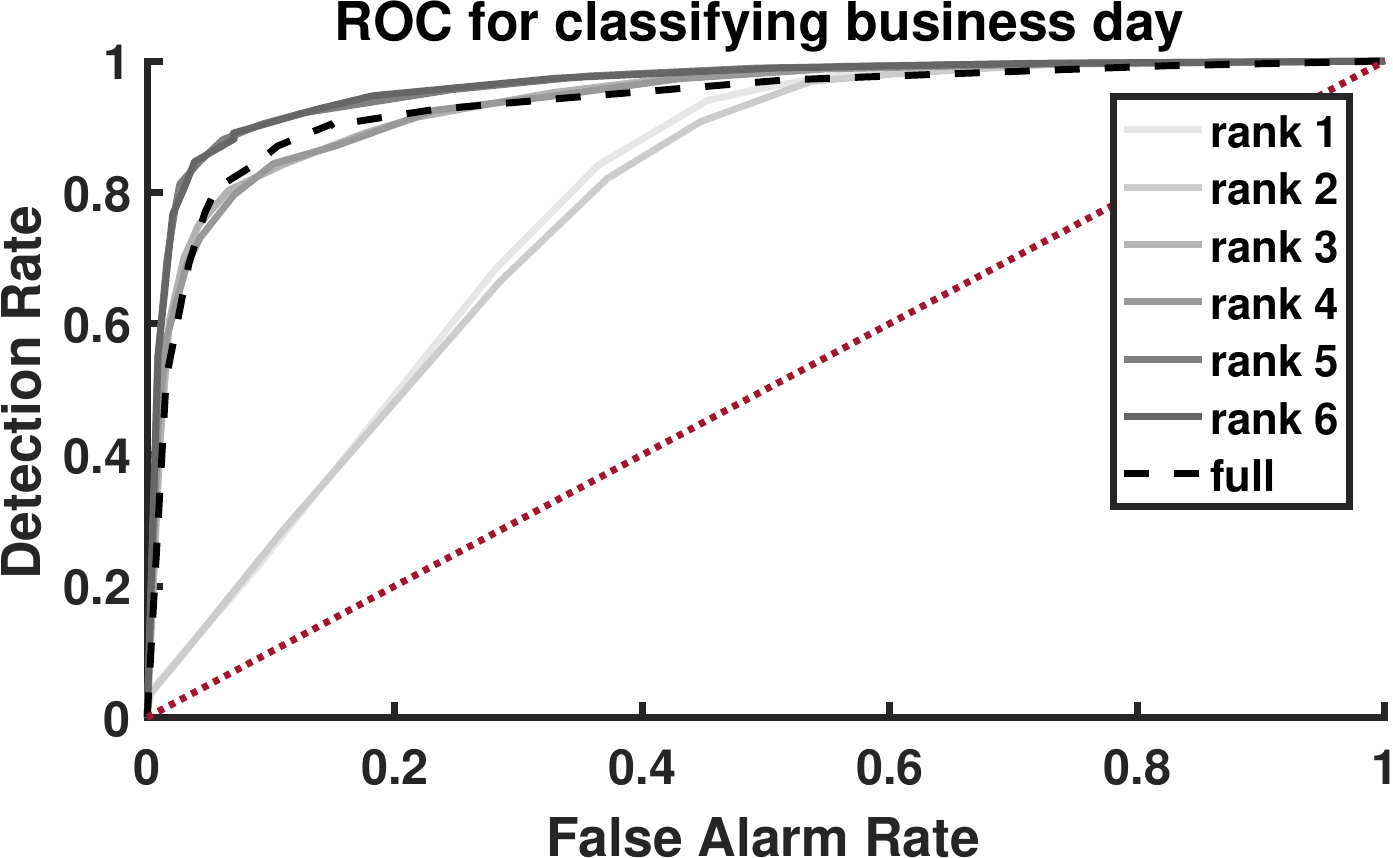}
	\caption{Receiver operating characteristics (ROCs) for classifying each day as a business day or non-business day, using the low-rank embedding provided by $\what{\L}$ learned from the SILVar model and using the full data}
	\label{fig:bikes_ROCs}		
\end{figure}
We also test the hypothesis that the learned $\what{\L}$ contains information about whether a day is business or non-business through its relatively low-dimensional effects on the data. We perform the singular value decomposition (SVD) on the optimally learned $\what{\L}=\what{\U}\what{\boldsymbol{\Sigma}}\what{\V}^\top$ for $n=360$ and project the data onto the $r$ top singular components $\wtil{\X}_r=\what{\boldsymbol{\Sigma}}_r\what{\V}_r^\top\X$. We then use $\wtil{\X}_r$ to train a linear support vector machine (SVM) to classify each day as either a business day or a non-business day, and compare the performance of this lower dimensional feature to that of using the full vector $\X$ to train a linear SVM. If our hypothesis is true then the performance of the classifier trained on $\wtil{\X}_r$ should be competitive with that of the classifier trained on $\X$. We use 50 training samples of $\wtil\X_r$ and of $\X$ and test on the remainder of the data. We repeat this 50 times by drawing a new batch of $50$ samples each time. We then vary the proportion of business to non-business days in the training sample to trace out a receiver operating characteristic (ROC).

In Figure~\ref{fig:bikes_ROCs}, we see the results of training linear SVM on $\wtil{\X}_r$ for $r\in\{1,...,6\}$ and on the full data for classifying business and non-business days. We see that using only the first two singular vectors, the performance is fairly poor. However, by simply taking 3 or 4 singular vectors, the classification performance almost matches that of the full data. Surprisingly, using the top 5 or 6 singular vectors achieves performance greater than that of the full data. This suggests that the projection may even play the role of a de-noising filter in some sense. Since we understand that behavior should correlate well with the day being business or non-business, this competitive performance of the classification using the lower dimensional features strongly suggests that the low-rank $\what\L$ indeed captures the effects of latent behavioral factors on the data.

\begin{figure}[t]	
	\includegraphics[width=\columnwidth]{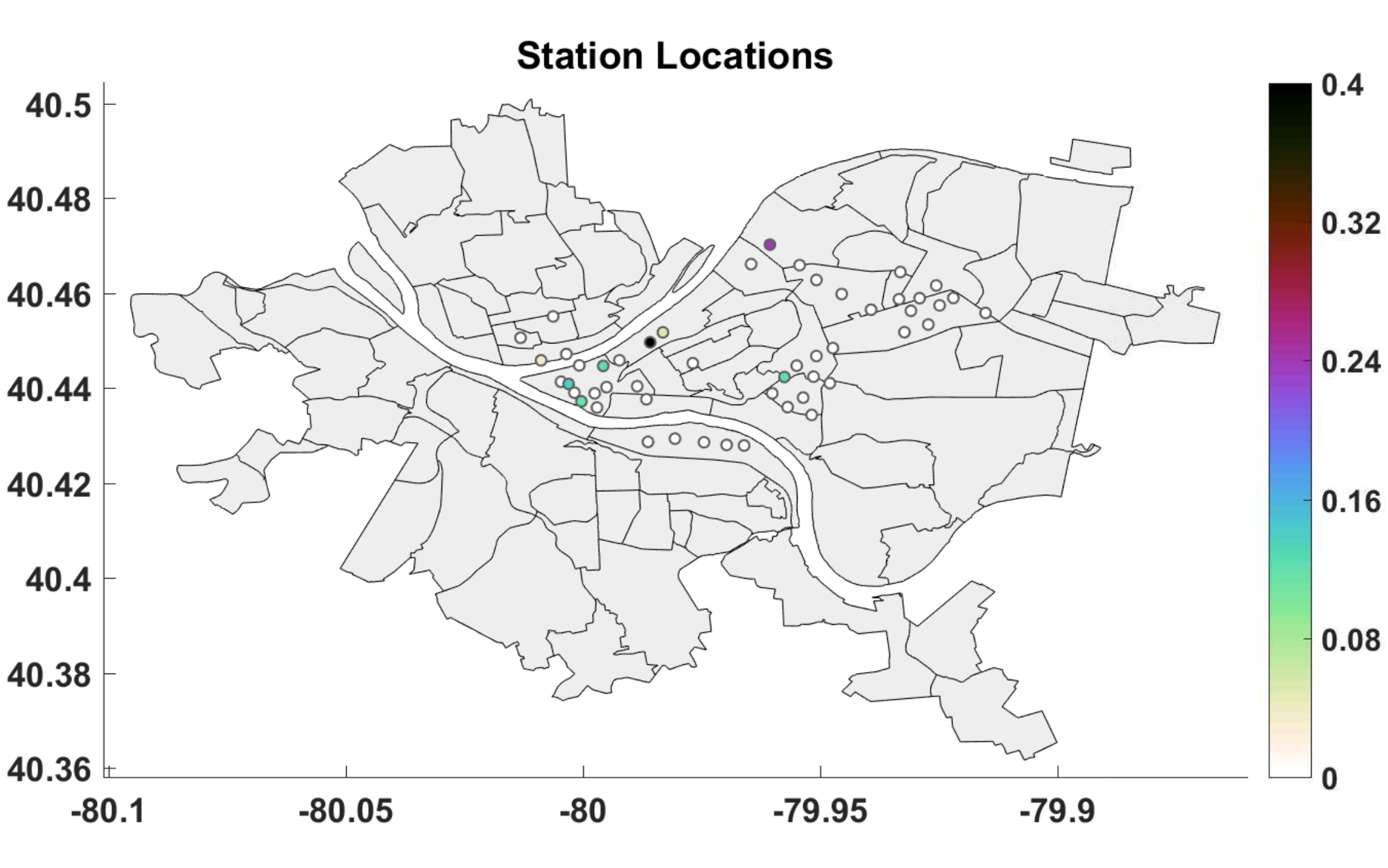}
	\caption{Intensities of the self-loop at each station}
	\label{fig:bikes_station_location}		
\end{figure}
Finally, in Figure~\ref{fig:bikes_station_location}, we plot the $(i,i)$ entries of the optimal $\what\A$ at $n=360$. This corresponds to locations for which incoming bike rides at 6:00AM-11:00AM are good predictors of outgoing bike rides at 11:00AM-2:00PM, beyond the effect of latent factors such as day of the week. We may intuitively expect this to correlate with locations that have restaurants open for lunch service, so that people would be likely to ride in for lunch or ride out after lunch. This is confirmed by observing that these stations are in Downtown (-80,40.44), the Strip District (-79.975, 40.45), Lawrenceville (-79.96, 40.47), and Oakland (-79.96, 40.44), known locations of many restaurants in Pittsburgh. It is especially interesting to note that Oakland, sandwiched between the University of Pittsburgh and Carnegie Mellon University, is included. Even though the target demographic is largely within walking distance, there is a high density of restaurants open for lunch, which may explain its non-zero coefficient. The remainder of the locations with non-zero coefficients $a_{ii}$ are also near high densities of lunch spots, while the other locations with coefficients $a_{ii}$ of zero are largely either near more residential areas or near restaurants more known for dinner or nightlife rather than lunch, such as Shadyside ($x\ge -79.95$) and Southside ($y\le 40.43$)).

\section{Conclusion}
\label{sec:conc}
Data exhibit complex dependencies, and it is often a challenge to deal with non-linearities and unmodeled effects when attempting to uncover meaningful relationships among various interacting entities that generate the data.
We introduce the SILVar model for performing semi-parametric sparse regression and estimating sparse graphs from data under the presence of non-linearities and latent factors or trends. The SILVar model estimates a non-linear link function $g$ as well as structured regression matrices $\A$ and $\L$ in a sparse and low-rank fashion. We justify the form of the model and relate it to existing methods for general PCA, multi-task regression, and vector autoregression. We provide computationally tractable algorithms for learning the SILVar model and demonstrate its performance against existing regression models and methods on both simulated and real data sets, namely 2011 US weather sensor network data and 2015-2016 Pittsburgh bike traffic data. We see from the simulated data that the SILVar model matches the performance of an Oracle GLM that knows the true link function and only needs to estimate $\A$ and $\L$; we show empirically on the temperature data that the learned $\L$ can capture the effects of underlying trends in time series while $\A$ represents a graph consistent with US weather patterns; and we see that, in the bike data, SILVar outperforms a GLM with a fixed link function, the learned $\L$ encodes latent behavioral aspects of the data, and $\A$ discovers notable locations consistent with the restaurant landscape of Pittsburgh.
%
%
%
%

\appendices

\section{Proof of Theorem~\ref{thm:equiv}}
\label{app:thm}
We first restate the theorem for convenience.
\setcounter{thm}{0}
\begin{thm}
	\label{thm:equiv_precise}
	Assume that $\what{g}'(0)\ne 0$ and that $|\what{g}''|\le J$ and $|\overline{g}''|\le J$ for some $J<\infty$. Furthermore, assume in models~\eqref{eq:SIM_MLE_full_obj} and~\eqref{eq:SIM_MLE_latent} that $\max_j \left(\|\what{\a}_j\|_1,\|\overline{\a}_j\|_1 +\|\overline{\b}_j\|_2 \right)\le k,$
	{\small \begin{equation*}
		\max\left( \Ebb[\|\x\|_2 \|\x\|_\infty^2], \Ebb[\|\x\|_2\|\x\|_\infty \|\z\|_2], \Ebb[\|\x\|_2 \|\z\|_2^2] \right) \le s_{Nr},
		\end{equation*}
	}where subscripts in the parameters on the RHS indicate that the bounds may grow with the values in the subscript.
	Then, the parameters $\what{\A}$ and $\overline{\A}$ from models~\eqref{eq:SIM_MLE_full_obj} and~\eqref{eq:SIM_MLE_latent} are related as
	\begin{equation*}
	\what{\A}= q(\overline{\A}+\L) + \E,
	\end{equation*}
	where $q=\frac{\overline{g}'(0)}{\what{g}'(0)}$, $\boldsymbol{\mu}_\x=\Ebb[\x_i]$,
	\begin{equation*}
	\begin{aligned}
	&\L=\Big(\overline{\B}\Ebb[\z\x^\top] + \left(\overline{\g}(\0)- \what{\g}(\0)\right)\boldsymbol{\mu}_\x^\top  \Big)\left(\Ebb[\x\x^\top]\right)^{\dagger}\\
	\Rightarrow&\textrm{rank}(\L)\le r+1,
	\end{aligned}
	\end{equation*}
	and $\E = \what{\A} - q(\overline{\A}+\L)$ is bounded as
	\begin{equation*}
	\!\begin{aligned}[t] \frac{1}{MN}\|\E\|_F & \le \! \frac{2J \sigma_\ell  \sqrt{N}}{\what{g}'(0)M}s_{Nr}k^2,
	\end{aligned}
	\end{equation*}
	where $\sigma_\ell=\left\|\left(\Ebb\left[\x\x^\top \right]\right)^\dagger \right\|_2$, the largest singular value of the pseudo-inverse of the covariance.
\end{thm}
~\\
\noindent\textit{Proof.} We begin with the Taylor series expansion, using the Lagrange form of the remainder,
{\small
	\begin{equation}
	\label{eq:proof_taylor}
	\begin{aligned}
	&\Ebb\left[ \left( \what{\g}(\0) \!+\! \what{g}'(0)\what{\A}\x \!-\! \overline{\g}(\0) \!-\! \overline{g}'(0)(\overline{\A}\x \!+\! \overline{\B}\z) \right) \x^\top \right]\\
	&\,=\!\Ebb\left[ \left( \what{\g}''(\boldsymbol{\xi}) \!\odot\! (\what{\A}\x_i).^\wedge 2 \!-\! \overline{\g}''(\boldsymbol{\eta})\odot(\overline{\A}\x_i \!+\! \overline{\B}\z).^\wedge 2 \right)\x^\top \right],
	\end{aligned}
	\end{equation}
}where 
\begin{equation*}
\begin{aligned}
&|\xi_j| \le |\what{\a}_j\x|, \qquad |\eta_j| \le |\overline{\a}_j\x+\overline{\b}_j\z|,
\end{aligned}
\end{equation*}
$\overline{\B}\!=\!\left(\overline{\b}_1\ldots\overline{\b}_m\right)^\top$ similarly to before, $\x\odot\y$ denotes the element-wise (Hadamard) product, and $\x.^\wedge 2=\x\odot\x$ denotes element-wise squaring.
First let us consider the quantity
\begin{align*}
(\overline{\a}_j\x \!+\! \overline{\b}_j\z)^2 &= (\overline{\a}_j\x)^2 \!+\! 2\overline{\a}_j\x\overline{\b}_j\z + (\overline{\b}_j\z)^2 \\
&\le (\|\overline{\a}_j\|_1 \|\x\|_\infty + \|\overline{\b}_j\|_2\|\z\|_2)^2
\end{align*}
Then, 
{\small
	\begin{align*}
	\|\!\left((\overline{\A}\x \!+\! \overline{\B}\z).\!^\wedge 2 \!\right)\! \x^{\!\top}\!\|_F \! & \!\le \|(\overline{\A}\x \!+\! \overline{\B}\z).\!^\wedge 2 \|_2 \|\x\|_2 \\
	& \le \|\x\|_2\sqrt{\sum_{j=1}^{N}(\overline{\a}_j\x \!+\! \overline{\b}_j\z)^4}\\
	&\overset{(a)}{\le} \|\x\|_2 \sum_{j=1}^{N}(\overline{\a}_j\x \!+\! \overline{\b}_j\z)^2\\
	&\le \! \|\x\|_2 \! \sum_{j=1}^{N}(\|\overline{\a}_j\|_1 \|\x \!\|_\infty \!+\! \|\overline{\b}_j\|_2\|\z\|_2)^2\\
	\Rightarrow \!\Ebb[\|\!\left((\overline{\A}\x \!+\! \overline{\B}\z).\!^\wedge 2 \!\right)\! \x^{\!\top}\!\|_F \!]&\le \! s_{Nr} \sum_{j=1}^{N}(\|\overline{\a}_j\|_1 \!+\! \|\overline{\b}_j\|_2)^2 \\
	&\le \! s_{Nr} Nk^2
	\end{align*}
}where $(a)$ follows from the fact that $\|\x\|_4 \le \|\x\|_2 \; \forall \x\in \Rbb^{N}$.

Similarly, we have
\begin{align*}
\Rightarrow \!\Ebb[\|\!((\what{\A}\x ).\!^\wedge 2 ) \x^{\!\top}\!\|_F \!]&\le \|\x\|_2\sum_{j=1}^{N}(\|\what{\a}_j\|_1 \|\x\|_\infty)^2\\
&\le \! s_{Nr} Nk^2
\end{align*}

To finish off, substituting the definition of $\E$ and $q$ into~\eqref{eq:proof_taylor} we have
{\small
	\begin{equation*}
	\!\begin{aligned}[t] \|\E\|_F & 
	\!\!\le\!\! J\Ebb\!\left[ \left\|\left(\! (\what{\A}\x_i).\!^\wedge 2 \!+\! (\overline{\A}\x_i \!+\! \overline{\B}\z_i).\!^\wedge 2 \!\right)\! \x_i^{\!\top} \! \right\|_F \!\right] \! \left\|\left( \what{g}'(0)\Ebb[\x_i\x_i^{\!\top} \! ] \! \right)^{\!\dagger} \right\|_F \\
	& \le \! \frac{2J \sigma_\ell  }{\what{g}'(0)}s_{Nr}k^2N\sqrt{N}.
	\end{aligned}
	\end{equation*}
}\hfill \qed

\section{Proof of Theorem~\ref{thm:perf}}
First, we present and prove some intermediate results. For convenience of notation, for the remainder of this section let $\boldsymbol{\Phi}=\what\A-\wtil\A$ and $\boldsymbol{\Psi}=\what\L-\wtil\L$.
\subsection{Propositions}
\setcounter{thm}{4}
\begin{prop}
	\label{prop:restricted}
	Under Assumption~\ref{as:Dev}, the solution $(\what\A,\what\L)$ to the optimization problem~\eqref{eq:SILVar_AR_opt} satisfies
	\begin{align*}
	&\gamma\|\boldsymbol{\Phi}_{S^c}\|_{1} \!+\! \|\boldsymbol{\Psi}_{R^c}\|_* \le 3( \gamma\|\boldsymbol{\Phi}_S\|_{1} \!+\!\|\boldsymbol{\Psi}_R\|_* )\\
	&\gamma\|\boldsymbol{\Phi}\|_{1} \!+\! \|\boldsymbol{\Psi}\|_* \le 4\sqrt{r_\L}( \tau\|\boldsymbol{\Phi}\|_{F} \!+\!\|\boldsymbol{\Psi}\|_F ).
	\end{align*}
	
\end{prop}
~\\
\noindent\textit{Proof.} We start with the convexity of the marginalized objective functional at $(\wtil\A,\wtil\L)$,
\begin{align*}
m(\what{\A} \!+\! \what{\L}) \!\ge\!  m(\wtil\A \!+\! \wtil\L) \!+\! \textrm{tr}\left(\left[\frac{1}{K}\what{\wtil{\boldsymbol{\Gamma}}}\X^\top \!\right]^\top \!(\what{\A} \!+\! \what{\L} \!-\! (\A \!+\! \L))\right)
\end{align*}

Then consider the optimality of the full objective functional at $(\what{g},\what{\A},\what{\L})$,
{\small
	\begin{align*}
	&m(\what{\A}+\what{\L}) +\lambda (\gamma\|\what{\A}\|_{1} + \|\what{\L}\|_* )\\
	&\quad = \what{F}_3(\Y,\X,\what{g},\what{\A}+\what{\L}) + (\gamma\|\what{\A}\|_{1} + \|\what{\L}\|_* ) \\
	&\quad \le \what{F}_3(\Y,\X,\what{\wtil{g}},\wtil{\A}+\wtil{\L})+(\gamma\|\wtil{\A}\|_{1} + \|\wtil{\L}\|_* )\\
	&\quad = m(\wtil{\A}+\wtil{\L}) + \lambda (\gamma\|\wtil{\A}\|_{1} + \|\wtil{\L}\|_* )\\
	\Rightarrow & m(\what{\A} \!+\! \what{\L}) \!-\! m(\wtil{\A}+\wtil{\L}) \le \lambda (\gamma\|\wtil{\A}\|_{1} \!+\! \|\wtil{\L}\|_* ) \!-\! \lambda (\gamma\|\what{\A}\|_{1} \!+\! \|\what{\L}\|_* )\\
	\Rightarrow &\textrm{tr}\left(\left[ \frac{1}{K}\what{\wtil{\boldsymbol{\Gamma}}}\X^\top \!\right]^\top \!  (\what{\A} \!+\! \what{\L} \!-\! (\wtil\A \!+\! \wtil\L))\right)\\
	& \quad \le \lambda (\gamma(\|\wtil\A\|_{1} \!-\! \|\what{\A}\|_{1}) \!+\!  \|\wtil\L\|_* \!-\! \|\what{\L}\|_* )
	\end{align*}
}where the last inequality utilizes convexity of the marginalized objective. Then using Assumption~\ref{as:Dev},

\begin{align*}
&-\!\frac{\lambda\gamma}{2}\|\boldsymbol{\Phi}\|_{1} \!-\! \frac{\lambda}{2}\|\boldsymbol{\Psi}\|_* \!\le\! \lambda\gamma(\|\wtil\A\|_{1} \!-\! \|\what{\A}\|_{1}) \!+\! \lambda(\|\wtil\L\|_* \!-\! \|\what{\L}\|_* )\\
\Rightarrow & 0 \le \frac{\gamma}{2}\|\boldsymbol{\Phi}\|_{1} \!+\! \gamma(\|\wtil\A\|_{1} \!-\! \|\what{\A}\|_{1}) \!+\! \frac{1}{2}\|\boldsymbol{\Psi}\|_* \!+\! (\|\L\|_* \!-\! \|\what{\L}\|_* )\\
\Rightarrow & 0 \le \frac{\gamma}{2}(\|\boldsymbol{\Phi}_S\|_{1} \!+\! \|\boldsymbol{\Phi}_{S^c}\|_{1}) \!+\! \gamma(\|\boldsymbol{\Phi}_S\|_{1} \!-\! \|\boldsymbol{\Phi}_{S^c}\|_{1})\\
& \qquad \!+\! \frac{1}{2}(\|\boldsymbol{\Psi}_R\|_* \!+\! \|\boldsymbol{\Psi}_{R^c}\|_*) \!+\! (\|\boldsymbol{\Psi}_R\|_* \!-\! \|\boldsymbol{\Psi}_{R^c}\|_* )\\
\Rightarrow& 0 \le -(\gamma\|\boldsymbol{\Phi}_{S^c}\|_{1} \!+\! \|\boldsymbol{\Psi}_{R^c}\|_*) \!+\! 3(\gamma\|\boldsymbol{\Phi}_S\|_{1} \!+\! \|\boldsymbol{\Psi}_R\|_* )
\end{align*}
where the penultimate inequality arises from decomposability of the norm. Specifically, 

\begin{align*}
\begin{cases}
\left| [\wtil\A]_s \right| - \left| [\what\A]_s \right| \le \left| [\wtil\A]_s - [\what\A]_s \right| = \left| [\boldsymbol{\Phi}]_s \right|, & s\in S\\
\left| [\wtil\A]_s \right| - \left| [\what\A]_s \right| = - \left| [\what\A]_s \right| = -\left| [\boldsymbol{\Phi}]_s \right|, & s\notin S
\end{cases}
\end{align*}
Thus, we have the first inequality
\begin{align*}
& \gamma\|\boldsymbol{\Phi}_{S^c}\|_{1} \!+\! \|\boldsymbol{\Psi}_{R^c}\|_* \le 3(\gamma\|\boldsymbol{\Phi}_S\|_{1} \!+\! \|\boldsymbol{\Psi}_R\|_* )
\end{align*}
Finally, for the second inequality,
\begin{align*}
\gamma\|\boldsymbol{\Phi}\|_{1} \!+\! \|\boldsymbol{\Psi}\|_* &\le 4( \gamma\|\boldsymbol{\Phi}_{S}\|_{1} \!+\! \|\boldsymbol{\Psi}_{R}\|_* )\\
&\le 4( \gamma\sqrt{s_\A}\|\boldsymbol{\Phi}_{S}\|_{F} \!+\! \sqrt{r_\L}\|\boldsymbol{\Psi}_{R}\|_F )\\
&\le 4\sqrt{r_\L}( \tau\|\boldsymbol{\Phi}_{S}\|_{F} \!+\!\|\boldsymbol{\Psi}_{R}\|_F )\\
&\le 4\sqrt{r_\L}( \tau\|\boldsymbol{\Phi}\|_{F} \!+\!\|\boldsymbol{\Psi}\|_F )
\end{align*}
\hfill \qed
\begin{prop}
	\label{prop:Inc}
	Under Assumptions~\ref{as:Dev} and \ref{as:Inc}, the solution of $(\what\A,\what\L)$ to the optimization problem~\eqref{eq:SILVar_AR_opt} satisfies
	\begin{align*} 
	2|\textrm{tr}(\boldsymbol{\Phi}^\top\boldsymbol{\Psi})| \le \mu(\gamma\|\boldsymbol{\Phi}\|_{1}+\|\boldsymbol{\Psi}\|_{*})^2.
	\end{align*}
\end{prop}
~\\
\noindent\textit{Proof.} This follows directly from Proposition~\ref{prop:restricted}, Assumption~\ref{as:Inc}, and applying Cauchy-Schwarz twice
\begin{align*}
2|\textrm{tr}(\boldsymbol{\Phi}^\top\boldsymbol{\Psi})| &\le \|\boldsymbol{\Psi}\|_{\infty}\|\boldsymbol{\Phi}\|_{1} +\|\boldsymbol{\Psi}\|_{*}\|\boldsymbol{\Phi}\|_{2}\\
&\le \mu\gamma(\gamma\|\boldsymbol{\Phi}\|_{1}+\|\boldsymbol{\Psi}\|_{*})\|\boldsymbol{\Phi}\|_1\\
&\quad +\mu(\gamma\|\boldsymbol{\Phi}\|_{1}+\|\boldsymbol{\Psi}\|_{*})\|\boldsymbol{\Psi}\|_*.
\end{align*}
\hfill \qed 
\subsection{Theorem}
Now we restate the theorem again for convenience.
\setcounter{thm}{2}
\begin{thm}
	Under Assumptions~\ref{as:RSC}-\ref{as:Inc}, the solution to the optimization problem~\eqref{eq:SILVar_AR_opt} satisfies
	\begin{align*}
	&\|\what\A-\wtil\A\|_F \le \frac{3\lambda\gamma\sqrt{s_\A}}{\alpha}\left(2 \!+\! \sqrt{\frac{2}{2 \!+\! \tau^2}}\right)\le\frac{9\lambda\gamma\sqrt{s_\A}}{\alpha}\\
	&\|\what\L-\wtil\L\|_F \le \frac{3\lambda\sqrt{r_\L}}{\alpha}\left(2 \!+\! \sqrt{\frac{2\tau^2}{1 \!+\! 2\tau^2}}\right)\le\frac{9\lambda\sqrt{r_\L}}{\alpha}.
	\end{align*}
\end{thm}
~\\
\noindent\textit{Proof.} Starting with Proposition~\ref{prop:restricted}, since our solution is in this restricted set, we can use the stronger convexity condition implied by Assumption~\ref{as:RSC},
\begin{align*}
m(\what{\A} \!+\! \what{\L}) \!\ge&   m(\wtil\A \!+\! \wtil\L) \!+\! \textrm{tr}\left(\left[\frac{1}{K}\what{\wtil{\boldsymbol{\Gamma}}}\X^\top \!\right]^\top \!\left(\what{\A} \!+\! \what{\L} \!-\! (\A \!+\! \L)\right)\right)\\
&\quad + \alpha \|\boldsymbol{\Phi}+\boldsymbol{\Psi}\|_F^2
\end{align*}

Revisiting the objective functional at optimality and skipping repetitive algebra (see proof for Proposition~\ref{prop:restricted}),
\begin{align*} 
\alpha \|\boldsymbol{\Phi} \!+\! \boldsymbol{\Psi}\|_F^2 \! &\le\! \frac{\lambda}{2}\left( 3(\gamma\|\boldsymbol{\Phi}_S\|_{1} \!+\! \|\boldsymbol{\Psi}_R\|_* ) \!-\!(\gamma\|\boldsymbol{\Phi}_{S^c}\|_{1} \!+\! \|\boldsymbol{\Psi}_{R^c}\|_*) \right)\\
&\le\! \frac{3\lambda}{2}(\gamma\|\boldsymbol{\Phi}\|_{1} \!+\! \|\boldsymbol{\Psi}\|_* )\\
&\le 6\lambda\sqrt{r_\L}(\tau\|\boldsymbol{\Phi}\|_{F} \!+\! \|\boldsymbol{\Psi}\|_F )
\end{align*} 

Now from Proposition~\ref{prop:Inc}, we have
\begin{align*} 
\|\boldsymbol{\Phi} \!+\! \boldsymbol{\Psi}\|_F^2 \! &\ge\!  \|\boldsymbol{\Phi}\|_F \!+\! \|\boldsymbol{\Psi}\|_F^2 - 2|\textrm{tr}(\boldsymbol{\Phi}^\top\boldsymbol{\Psi})| \\
&\ge \|\boldsymbol{\Phi}\|_F \!+\! \|\boldsymbol{\Psi}\|_F^2 - \mu(\gamma\|\boldsymbol{\Phi}\|_{1}+\|\boldsymbol{\Psi}\|_{*})^2
\end{align*} 

Combining the previous two inequalities, we have
{\small
	\begin{align*} 
	&\alpha (\|\boldsymbol{\Phi}\|_F^2 \!+\! \|\boldsymbol{\Psi}\|_F^2) \!-\! \alpha\mu(\gamma\|\boldsymbol{\Phi}\|_{1} \!+\!\|\boldsymbol{\Psi}\|_{*})^2  \! \le\! 6\lambda\sqrt{r_\L}(\tau\|\boldsymbol{\Phi}\|_{F} \!+\! \|\boldsymbol{\Psi}\|_F )\\
	\Rightarrow& \alpha (\|\boldsymbol{\Phi}\|_F^2 \!+\! \|\boldsymbol{\Psi}\|_F^2) \!-\! 16\alpha\mu r_\L (\tau\|\boldsymbol{\Phi}\|_{F} \!+\!\|\boldsymbol{\Psi}\|_{F})^2  \\
	&\qquad \le\! 6\lambda\sqrt{r_\L}(\tau\|\boldsymbol{\Phi}\|_{F} \!+\! \|\boldsymbol{\Psi}\|_F )\\
	\Rightarrow& \boldsymbol{\zeta}^\top\Q\boldsymbol{\zeta}\le 0
	\end{align*}
}where
\begin{align*} 
&\boldsymbol{\zeta}= ( \|\boldsymbol{\Phi}\|_F \quad \|\boldsymbol{\Psi}\|_F \quad 1)^\top\\
&\Q=\left(\begin{array}{ccc}
\alpha\left(\frac{2+\tau^2}{2(1+\tau^2)}\right) & -\alpha\left(\frac{\tau}{2(1+\tau^2)}\right) & -3\lambda'\tau \\
-\alpha\left(\frac{\tau}{2(1+\tau^2)}\right) & \alpha\left(\frac{1+2\tau^2}{2(1+\tau^2)}\right) & -3\lambda' \\
-3\lambda'\tau & -3\lambda' & 0
\end{array}\right),
\end{align*}
which is a conic in standard form, $\lambda'=\lambda\sqrt{r_\L}$, and the entries of $\Q$ follow from taking $\mu$ given in Assumption~\ref{as:Inc}.

Checking its discriminant,
\begin{align*}
\left(2+\tau^2\right)(1+2\tau^2) - \tau^2 > 0
\end{align*}
so we have that the conic equation describes an ellipse, and there are bounds on the values of $\|\boldsymbol{\Phi}\|_F$ and  $\|\boldsymbol{\Psi}\|_F$.

For these individual bounds, we consider the points at which the gradients of $\boldsymbol{\zeta}^\top\Q\boldsymbol{\zeta}$ vanish w.r.t. each of $\|\boldsymbol{\Phi}\|_F$ and  $\|\boldsymbol{\Psi}\|_F$. For $\|\boldsymbol{\Phi}\|_F$,
\begin{align*}
&\partial_{\|\boldsymbol{\Phi}\|_F}\boldsymbol{\zeta}^\top\Q\boldsymbol{\zeta}=0 \\
\Rightarrow &
\alpha\left(\frac{2+\tau^2}{2(1+\tau^2)}\right)\|\boldsymbol{\Phi}\|_F ^* = 3\lambda'\tau + \alpha\left(\frac{\tau}{2(1+\tau^2)}\right) \|\boldsymbol{\Psi}\|_F^*.
\end{align*}

Plugging this into the equation defined by $\boldsymbol{\zeta}^\top\Q\boldsymbol{\zeta}=0$ yields
\begin{align*}
\|\boldsymbol{\Phi}\|_F^* = \frac{3\lambda'\tau}{\alpha}\left(2\pm\sqrt{\frac{2}{2+\tau^2}}\right).
\end{align*}
Since we are seeking an upper bound for $\|\boldsymbol{\Phi}\|_F$, it can be seen that we take the positive root,

\begin{align*}
\|\boldsymbol{\Phi}\|_F \le  \frac{3\lambda'\tau}{\alpha}\left(2+\sqrt{\frac{2}{2+\tau^2}}\right) \le\frac{9\lambda'\tau}{\alpha}
\end{align*}

Similarly, for $\|\boldsymbol{\Psi}\|_F$,
\begin{align*}
&\partial_{\|\boldsymbol{\Psi}\|_F}\boldsymbol{\zeta}^\top\Q\boldsymbol{\zeta}=0 \\
\Rightarrow &
\alpha\left(\frac{2+\tau^2}{2(1+\tau^2)}\right)\|\boldsymbol{\Psi}\|_F^* = 3\lambda'\tau + \alpha\left(\frac{\tau}{2(1+\tau^2)}\right) \|\boldsymbol{\Phi}\|_F^*.
\end{align*}

Finally, plugging this into $\boldsymbol{\zeta}^\top\Q\boldsymbol{\zeta}=0$ and solving for the upper bound yields
\begin{align*}
\|\boldsymbol{\Psi}\|_F \le \frac{3\lambda'}{\alpha}\left(2+\sqrt{\frac{2\tau^2}{1+2\tau^2}}\right)\le\frac{9\lambda'}{\alpha}
\end{align*}
\hfill \qed

\ifCLASSOPTIONcaptionsoff
\newpage
\fi

\bibliographystyle{IEEEtran}
\bibliography{jmei}

\begin{thebibliography}{10}
\providecommand{\url}[1]{#1}
\csname url@samestyle\endcsname
\providecommand{\newblock}{\relax}
\providecommand{\bibinfo}[2]{#2}
\providecommand{\BIBentrySTDinterwordspacing}{\spaceskip=0pt\relax}
\providecommand{\BIBentryALTinterwordstretchfactor}{4}
\providecommand{\BIBentryALTinterwordspacing}{\spaceskip=\fontdimen2\font plus
\BIBentryALTinterwordstretchfactor\fontdimen3\font minus
  \fontdimen4\font\relax}
\providecommand{\BIBforeignlanguage}[2]{{%
\expandafter\ifx\csname l@#1\endcsname\relax
\typeout{** WARNING: IEEEtran.bst: No hyphenation pattern has been}%
\typeout{** loaded for the language `#1'. Using the pattern for}%
\typeout{** the default language instead.}%
\else
\language=\csname l@#1\endcsname
\fi
#2}}
\providecommand{\BIBdecl}{\relax}
\BIBdecl

\bibitem{tibshirani_regression_1996}
\BIBentryALTinterwordspacing
R.~Tibshirani, ``Regression {Shrinkage} and {Selection} via the {Lasso},''
  \emph{Journal of the Royal Statistical Society. Series B (Methodological)},
  vol.~58, no.~1, pp. 267--288, 1996. [Online]. Available:
  \url{http://www.jstor.org/stable/2346178}
\BIBentrySTDinterwordspacing

\bibitem{friedman_sparse_2008}
\BIBentryALTinterwordspacing
J.~Friedman, T.~Hastie, and R.~Tibshirani, ``Sparse inverse covariance
  estimation with the graphical {Lasso},'' \emph{Biostatistics}, vol.~9, no.~3,
  pp. 432--41, Jul. 2008. [Online]. Available:
  \url{http://www.pubmedcentral.nih.gov/articlerender.fcgi?artid=3019769&tool=pmcentrez&rendertype=abstract}
\BIBentrySTDinterwordspacing

\bibitem{bolstad_causal_2011}
A.~Bolstad, B.~Van~Veen, and R.~Nowak, ``Causal network inference via group
  sparse regularization,'' \emph{IEEE Transactions on Signal Processing},
  vol.~59, no.~6, pp. 2628--2641, Jun. 2011.

\bibitem{basu_regularized_2015}
\BIBentryALTinterwordspacing
S.~Basu and G.~Michailidis, ``\BIBforeignlanguage{EN}{Regularized estimation in
  sparse high-dimensional time series models},''
  \emph{\BIBforeignlanguage{EN}{The Annals of Statistics}}, vol.~43, no.~4, pp.
  1535--1567, Aug. 2015. [Online]. Available:
  \url{http://projecteuclid.org/euclid.aos/1434546214}
\BIBentrySTDinterwordspacing

\bibitem{granger_investigating_1969}
\BIBentryALTinterwordspacing
C.~W.~J. Granger, ``Investigating causal relations by econometric models and
  cross-spectral methods,'' \emph{Econometrica}, vol.~37, no.~3, pp. 424--438,
  Aug. 1969. [Online]. Available: \url{http://www.jstor.org/stable/1912791}
\BIBentrySTDinterwordspacing

\bibitem{sandryhaila_discrete_2013}
A.~Sandryhaila and J.~M.~F. Moura, ``Discrete signal processing on graphs,''
  \emph{IEEE Transactions on Signal Processing}, vol.~61, no.~7, pp.
  1644--1656, Apr. 2013.

\bibitem{sandryhaila_discrete_2014}
------, ``Discrete signal processing on graphs: {Frequency} analysis,''
  \emph{IEEE Transactions on Signal Processing}, vol.~62, no.~12, pp.
  3042--3054, Jun. 2014.

\bibitem{mei_signal_2017}
J.~Mei and J.~M.~F. Moura, ``Signal processing on graphs: causal modeling of
  unstructured data,'' \emph{IEEE Transactions on Signal Processing}, vol.~65,
  no.~8, pp. 2077--2092, Apr. 2017.

\bibitem{chandrasekaran_latent_2012}
\BIBentryALTinterwordspacing
V.~Chandrasekaran, P.~A. Parrilo, and A.~S. Willsky,
  ``\BIBforeignlanguage{EN}{Latent variable graphical model selection via
  convex optimization},'' \emph{\BIBforeignlanguage{EN}{The Annals of
  Statistics}}, vol.~40, no.~4, pp. 1935--1967, Aug. 2012. [Online]. Available:
  \url{http://projecteuclid.org/euclid.aos/1351602527}
\BIBentrySTDinterwordspacing

\bibitem{jalali_learning_2011}
\BIBentryALTinterwordspacing
A.~Jalali and S.~Sanghavi, ``Learning the dependence graph of time series with
  latent factors,'' \emph{arXiv:1106.1887 [cs]}, Jun. 2011, arXiv: 1106.1887.
  [Online]. Available: \url{http://arxiv.org/abs/1106.1887}
\BIBentrySTDinterwordspacing

\bibitem{bahadori_fast_2013}
\BIBentryALTinterwordspacing
M.~T. Bahadori, Y.~Liu, and E.~P. Xing, ``Fast structure learning in
  generalized stochastic processes with latent factors,'' in \emph{Proceedings
  of the 19th {ACM} {SIGKDD} {International} {Conference} on {Knowledge}
  {Discovery} and {Data} {Mining}}, ser. {KDD} '13.\hskip 1em plus 0.5em minus
  0.4em\relax New York, NY, USA: ACM, 2013, pp. 284--292. [Online]. Available:
  \url{http://doi.acm.org/10.1145/2487575.2487578}
\BIBentrySTDinterwordspacing

\bibitem{ganti_learning_2015}
\BIBentryALTinterwordspacing
R.~Ganti, N.~Rao, R.~M. Willett, and R.~Nowak, ``Learning single index models
  in high dimensions,'' \emph{arXiv:1506.08910 [cs, stat]}, Jun. 2015, arXiv:
  1506.08910. [Online]. Available: \url{http://arxiv.org/abs/1506.08910}
\BIBentrySTDinterwordspacing

\bibitem{bregman_relaxation_1967}
\BIBentryALTinterwordspacing
L.~M. Bregman, ``The relaxation method of finding the common point of convex
  sets and its application to the solution of problems in convex programming,''
  \emph{USSR Computational Mathematics and Mathematical Physics}, vol.~7,
  no.~3, pp. 200--217, Jan. 1967. [Online]. Available:
  \url{http://www.sciencedirect.com/science/article/pii/0041555367900407}
\BIBentrySTDinterwordspacing

\bibitem{banerjee_clustering_2005}
\BIBentryALTinterwordspacing
A.~Banerjee, S.~Merugu, I.~S. Dhillon, and J.~Ghosh, ``Clustering with
  {Bregman} divergences,'' \emph{Journal of Machine Learning Research}, vol.~6,
  no. Oct, pp. 1705--1749, 2005. [Online]. Available:
  \url{http://www.jmlr.org/papers/v6/banerjee05b.html}
\BIBentrySTDinterwordspacing

\bibitem{acharyya_parameter_2015}
\BIBentryALTinterwordspacing
S.~Acharyya and J.~Ghosh, ``Parameter estimation of {Generalized} {Linear}
  {Models} without assuming their link function,'' in \emph{Proceedings of the
  {Eighteenth} {International} {Conference} on {Artificial} {Intelligence} and
  {Statistics}}, 2015, pp. 10--18. [Online]. Available:
  \url{http://www.jmlr.org/proceedings/papers/v38/acharyya15.html}
\BIBentrySTDinterwordspacing

\bibitem{ichimura_semiparametric_1993}
\BIBentryALTinterwordspacing
H.~Ichimura, ``Semiparametric {Least} {Squares} ({SLS}) and weighted {SLS}
  estimation of {Single}-{Index} {Models},'' \emph{Journal of Econometrics},
  vol.~58, no.~1, pp. 71--120, Jul. 1993. [Online]. Available:
  \url{http://www.sciencedirect.com/science/article/pii/030440769390114K}
\BIBentrySTDinterwordspacing

\bibitem{candes_robust_2011}
\BIBentryALTinterwordspacing
E.~J. Cand\`{e}s, X.~Li, Y.~Ma, and J.~Wright, ``Robust {Principal} {Component}
  {Analysis}?'' \emph{J. ACM}, vol.~58, no.~3, pp. 11:1--11:37, Jun. 2011.
  [Online]. Available: \url{http://doi.acm.org/10.1145/1970392.1970395}
\BIBentrySTDinterwordspacing

\bibitem{ganti_matrix_2015}
\BIBentryALTinterwordspacing
R.~S. Ganti, L.~Balzano, and R.~Willett, ``Matrix completion under monotonic
  {Single} {Index} {Models},'' in \emph{Advances in {Neural} {Information}
  {Processing} {Systems} 28}, C.~Cortes, N.~D. Lawrence, D.~D. Lee,
  M.~Sugiyama, and R.~Garnett, Eds.\hskip 1em plus 0.5em minus 0.4em\relax
  Curran Associates, Inc., 2015, pp. 1873--1881. [Online]. Available:
  \url{http://papers.nips.cc/paper/5916-matrix-completion-under-monotonic-single-index-models.pdf}
\BIBentrySTDinterwordspacing

\bibitem{udell_generalized_2016}
\BIBentryALTinterwordspacing
M.~Udell, C.~Horn, R.~Zadeh, and S.~Boyd,
  ``\BIBforeignlanguage{English}{Generalized {Low} {Rank} {Models}},''
  \emph{\BIBforeignlanguage{English}{Foundations and Trends® in Machine
  Learning}}, vol.~9, no.~1, pp. 1--118, Jun. 2016. [Online]. Available:
  \url{http://ftp.nowpublishers.com/article/Details/MAL-055}
\BIBentrySTDinterwordspacing

\bibitem{granger_causality_1988}
\BIBentryALTinterwordspacing
C.~W.~J. Granger, ``Causality, cointegration, and control,'' \emph{J. of Econ.
  Dynamics and Control}, vol.~12, no. 2–3, pp. 551--559, Jun. 1988. [Online].
  Available:
  \url{http://www.sciencedirect.com/science/article/pii/0165188988900553}
\BIBentrySTDinterwordspacing

\bibitem{mei_signal_2015-1}
J.~Mei and J.~M.~F. Moura, ``Signal processing on graphs: {Estimating} the
  structure of a graph,'' in \emph{2015 {IEEE} {International} {Conference} on
  {Acoustics}, {Speech} and {Signal} {Processing} ({ICASSP})}, Apr. 2015, pp.
  5495--5499.

\bibitem{dykstra_isotonic_1981}
\BIBentryALTinterwordspacing
R.~L. Dykstra, ``An isotonic regression algorithm,'' \emph{Journal of
  Statistical Planning and Inference}, vol.~5, no.~4, pp. 355--363, Jan. 1981.
  [Online]. Available:
  \url{http://www.sciencedirect.com/science/article/pii/0378375881900367}
\BIBentrySTDinterwordspacing

\bibitem{lopez_acceleration_2016}
\BIBentryALTinterwordspacing
W.~L\'{o}pez and M.~Raydan, ``\BIBforeignlanguage{en}{An acceleration scheme
  for {Dykstra}’s algorithm},'' \emph{\BIBforeignlanguage{en}{Computational
  Optimization and Applications}}, vol.~63, no.~1, pp. 29--44, Jan. 2016.
  [Online]. Available:
  \url{https://link.springer.com/article/10.1007/s10589-015-9768-y}
\BIBentrySTDinterwordspacing

\bibitem{yeganova_isotonic_2009}
\BIBentryALTinterwordspacing
L.~Yeganova and W.~J. Wilbur, ``\BIBforeignlanguage{en}{Isotonic {Regression}
  under {Lipschitz} {Constraint}},'' \emph{\BIBforeignlanguage{en}{Journal of
  Optimization Theory and Applications}}, vol. 141, no.~2, pp. 429--443, May
  2009. [Online]. Available:
  \url{https://link.springer.com/article/10.1007/s10957-008-9477-0}
\BIBentrySTDinterwordspacing

\bibitem{kakade_efficient_2011}
\BIBentryALTinterwordspacing
S.~Kakade, A.~T. Kalai, V.~Kanade, and O.~Shamir, ``Efficient {Learning} of
  {Generalized} {Linear} and {Single} {Index} {Models} with {Isotonic}
  {Regression},'' \emph{arXiv:1104.2018 [cs, stat]}, Apr. 2011, arXiv:
  1104.2018. [Online]. Available: \url{http://arxiv.org/abs/1104.2018}
\BIBentrySTDinterwordspacing

\bibitem{odonoghue_adaptive_2013}
\BIBentryALTinterwordspacing
B.~O'Donoghue and E.~Cand\`{e}s, ``\BIBforeignlanguage{en}{Adaptive restart for
  accelerated gradient schemes},'' \emph{\BIBforeignlanguage{en}{Foundations of
  Computational Mathematics}}, vol.~15, no.~3, pp. 715--732, Jul. 2013.
  [Online]. Available:
  \url{http://link.springer.com/article/10.1007/s10208-013-9150-3}
\BIBentrySTDinterwordspacing

\bibitem{schmidt_optimizing_2009}
M.~Schmidt, E.~V.~D. Berg, M.~P. Friedl, and K.~Murphy, ``Optimizing costly
  functions with simple constraints: {A} limited-memory projected
  quasi-{Newton} algorithm,'' in \emph{Proc. of {Conf}. on {Artificial}
  {Intelligence} and {Statistics}}, 2009, pp. 456--463.

\bibitem{nocedal_numerical_1999}
J.~Nocedal and S.~J. Wright, \emph{Numerical {Optimization}}.\hskip 1em plus
  0.5em minus 0.4em\relax Springer, 1999.

\bibitem{lucet_faster_1997}
\BIBentryALTinterwordspacing
Y.~Lucet, ``\BIBforeignlanguage{en}{Faster than the {Fast} {Legendre}
  {Transform}, the {Linear}-time {Legendre} {Transform}},''
  \emph{\BIBforeignlanguage{en}{Numerical Algorithms}}, vol.~16, no.~2, pp.
  171--185, Mar. 1997. [Online]. Available:
  \url{https://link.springer.com/article/10.1023/A:1019191114493}
\BIBentrySTDinterwordspacing

\bibitem{negahban_unified_2012}
\BIBentryALTinterwordspacing
S.~N. Negahban, P.~Ravikumar, M.~J. Wainwright, and B.~Yu,
  ``\BIBforeignlanguage{EN}{A unified framework for high-dimensional analysis
  of {M}-estimators with decomposable regularizers},''
  \emph{\BIBforeignlanguage{EN}{Statistical Science}}, vol.~27, no.~4, pp.
  538--557, Nov. 2012. [Online]. Available:
  \url{http://projecteuclid.org/euclid.ss/1356098555}
\BIBentrySTDinterwordspacing

\bibitem{chandrasekaran_rank-sparsity_2011}
\BIBentryALTinterwordspacing
V.~Chandrasekaran, S.~Sanghavi, P.~Parrilo, and A.~Willsky, ``Rank-sparsity
  incoherence for matrix decomposition,'' \emph{SIAM Journal on Optimization},
  vol.~21, no.~2, pp. 572--596, Apr. 2011. [Online]. Available:
  \url{http://epubs.siam.org/doi/abs/10.1137/090761793}
\BIBentrySTDinterwordspacing

\bibitem{candes_matrix_2010}
E.~Cand\`{e}s and Y.~Plan, ``Matrix completion with noise,'' \emph{Proceedings
  of the IEEE}, vol.~98, no.~6, pp. 925--936, Jun. 2010.

\bibitem{hedge_summary_2011}
\BIBentryALTinterwordspacing
C.~Hedge, \emph{Summary of {February} 1-3, 2011 {Central} and {Eastern}
  {U}.{S}. {Winter} {Storm}}.\hskip 1em plus 0.5em minus 0.4em\relax Weather
  Prediction Center, National Oceanic and Atmospheric Administration, Feb.
  2011. [Online]. Available:
  \url{http://www.wpc.ncep.noaa.gov/winter_storm_summaries/event_reviews/2011/Feb1-3_Central_Eastern_Winterstorm.pdf}
\BIBentrySTDinterwordspacing

\bibitem{hamrick_mid-atlantic_2011}
\BIBentryALTinterwordspacing
D.~Hamrick, \emph{Mid-{Atlantic} and {Northeast} {U}.{S}. {Winter} {Storm}
  {January} 26-27, 2011}.\hskip 1em plus 0.5em minus 0.4em\relax Weather
  Prediction Center, National Oceanic and Atmospheric Administration, Jan.
  2011. [Online]. Available:
  \url{http://www.wpc.ncep.noaa.gov/winter_storm_summaries/event_reviews/2011/Mid-Atlantic_Northeast_WinterStorm_Jan2011.pdf}
\BIBentrySTDinterwordspacing

\bibitem{noauthor_healthy_2016}
\BIBentryALTinterwordspacing
``Healthy {Ride} {Pittsburgh},'' https://healthyridepgh.com/data/, Oct. 2016.
  [Online]. Available: \url{https://healthyridepgh.com/data/}
\BIBentrySTDinterwordspacing

\end{thebibliography}

\balance
\begin{IEEEbiography}[{\includegraphics[width=1in,height=1.25in,clip,keepaspectratio]{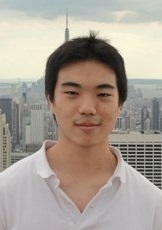}}]{Jonathan Mei}
	received his B.S. and M.Eng. degrees in electrical engineering both from the Massachusetts Institute of Technology in Cambridge, MA in 2013. He is currently pursuing his Ph.D. degree in electrical and computer engineering at Carnegie Mellon University in Pittsburgh, PA. His research interests include computational photography, image processing, reinforcement learning, graph signal processing, non-stationary time series analysis, and high-dimensional optimization.
\end{IEEEbiography}%
\begin{IEEEbiography}[{\includegraphics[width=1in,height=1.25in,clip,keepaspectratio]{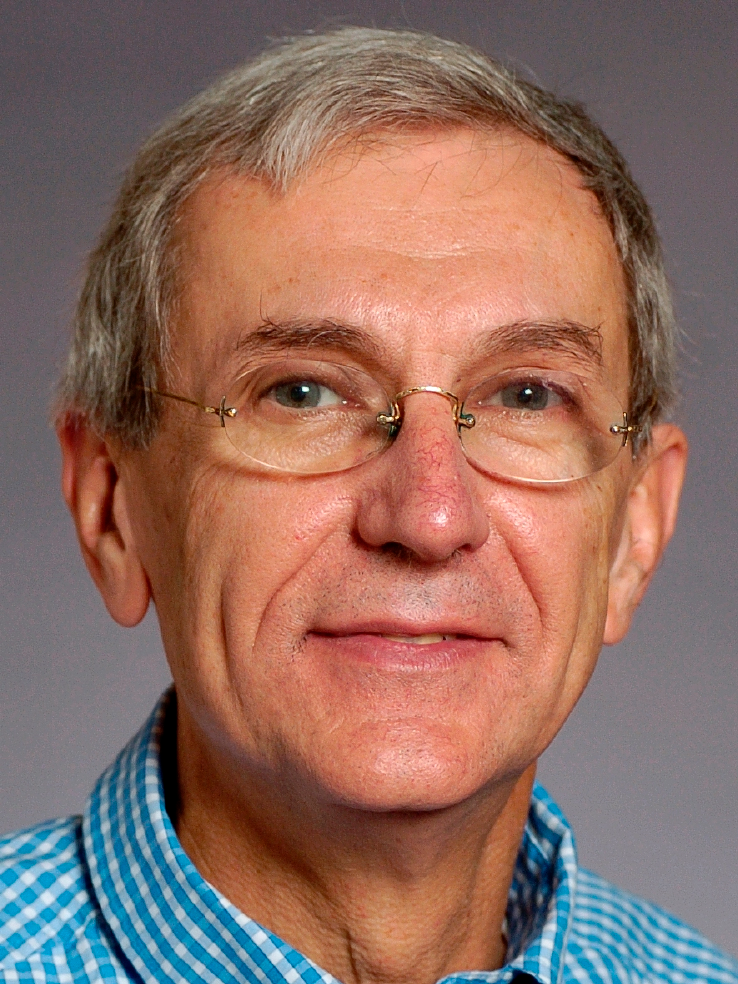}}]{Jos\'e M.~F.~Moura}(S'71--M'75--SM'90--F'94) is the Philip L.~and Marsha Dowd University Professor at Carnegie Mellon University (CMU). He holds the engenheiro electrot\'{e}cnico degree from Instituto Superior T\'ecnico (IST), Lisbon, Portugal, and the M.Sc., E.E., and D.Sc.~degrees in EECS from the Massachusetts Institute of Technology (MIT), Cambridge, MA. He was on the faculty at IST and has held visiting faculty appointments at MIT and New York University (NYU). He founded and directs a large education and research program between CMU and Portugal, www.icti.cmu.edu.
	
	His research interests are on  data science and graph signal processing. He has published over 550 papers and holds 14~US patents. Two of his patents (co-inventor A. Kav$\check{\textrm{c}}$i\'c) are used in over three billion disk drives in 60~\% of all computers sold in the last 13 years worldwide and were, in 2016, the subject of a 750 million dollar settlement between CMU and a chip manufacturer, the largest ever university verdict/settlement in the information technologies area.
	
	Dr. Moura is the 2018 IEEE President Elect, was IEEE Technical Activities Vice-President, IEEE Division IX Director, President of the IEEE Signal Processing Society(SPS), and Editor in Chief for the IEEE Transactions in Signal Processing.
	
	Dr. Moura received the Technical Achievement Award and the Society Award from the IEEE Signal Processing Society, and the CMU College of Engineering Distinguished Professor of Engineering Award. He is a Fellow of the IEEE, the American Association for the Advancement of Science (AAAS), a corresponding member of the Academy of Sciences of Portugal, Fellow of the US National Academy of Inventors, and a member of the US National Academy of Engineering.
	
\end{IEEEbiography} 

\end{document}